\documentclass[acmsmall]{acmart}
\usepackage[T1]{fontenc}
\usepackage{adjustbox}
\usepackage{threeparttable}
\usepackage{multirow}
\usepackage{tabularx}
\usepackage{enumitem}
\usepackage{pifont}
\usepackage{caption}
\usepackage{graphicx}
\usepackage{float} 
\usepackage{subcaption}
\usepackage{makecell}
\usepackage{soul}
\usepackage{tcolorbox}
\usepackage{mdframed}
\tcbuselibrary{skins}

\usepackage{caption}
\usepackage{graphicx}
\usepackage{float} 
\usepackage{subcaption}
\usepackage{soul}
\usepackage{mdframed}
\usepackage{makecell} 
\usepackage[normalem]{ulem} 
\usepackage[table]{xcolor}

\newcommand{\tech}{\textsc{EMBERS}}

\AtBeginDocument{%
  }

\setcopyright{acmlicensed}
\copyrightyear{2025}
\acmYear{2025}
\acmDOI{XXXXXXX.XXXXXXX}
\acmConference[Conference acronym 'XX]{Make sure to enter the correct
  conference title from your rights confirmation email}{June 03--05,
  2018}{Woodstock, NY}
\acmISBN{978-1-4503-XXXX-X/2018/06}
\newcounter{finding}
\newcommand{\finding}[1]{\refstepcounter{finding}
    \vspace{6pt}
    \begin{mdframed}[linecolor=gray,roundcorner=12pt,backgroundcolor=gray!15,linewidth=3pt,innerleftmargin=2pt, leftmargin=0cm,rightmargin=0cm,topline=false,bottomline=false,rightline = false]
        #1
    \end{mdframed}
    \vspace{4pt}
}

\begin{document}

\title{EMBERS: \underline{E}mpirically \underline{M}otivated Ro\underline{b}ustness Testing of D\underline{e}ep \underline{R}esearch \underline{S}ystems via Defect-Guided Fuzzing}

\author{Shuochuan Li}
\email{lishuochuan@tju.edu.cn}
\affiliation{%
  \institution{Tianjin University}
  \city{Tianjin}
  \country{China}
}

\author{Zan Wang}
\authornote{Corresponding Author.}
\email{wangzan@tju.edu.cn}
\affiliation{%
  \institution{Tianjin University}
  \city{Tianjin}
  \country{China}
}

\author{Dong Wang}
\affiliation{%
  \institution{Tianjin University}
  \city{Tianjin}
  \country{China}
}

\author{Yuhang Jiang}
\affiliation{%
  \institution{Tianjin University}
  \city{Tianjin}
  \country{China}
}

\author{Huajin Cui}
\affiliation{%
  \institution{Tianjin University}
  \city{Tianjin}
  \country{China}
}

\author{Chen Zhang}
\affiliation{%
  \institution{Huawei Technologies Co., Ltd.}
  \city{Shenzhen}
  \country{China}
}

\author{Gang Wang}
\affiliation{%
  \institution{Huawei Technologies Co., Ltd.}
  \city{Shenzhen}
  \country{China}
}

\author{Junjie Chen}
\affiliation{%
  \institution{Tianjin University}
  \city{Tianjin}
  \country{China}
}

\renewcommand{\shortauthors}{Li et al.}

\begin{abstract}
Deep Research (DR) systems extend large language models into autonomous agents that decompose complex information needs, iteratively search the web, and synthesize evidence to answer challenging questions.
However, their multi-stage execution may introduce robustness risks, as variations in user questions or intermediate search inputs may propagate across research stages and lead to incorrect answers.
Existing evaluations focus on task-solving capabilities or robustness against external evidence disturbances, leaving intrinsic weaknesses in DR execution insufficiently explored.
Thus, we conduct an empirical study of 1,932 execution trajectories from three representative DR systems and identify nine recurring defect categories across subquestion formulation, online search, and final reasoning.
Guided by these findings, we propose \tech{}, an empirically motivated, defect-guided fuzzing framework for robustness testing of DR systems.
Starting from correctly answered seed questions, \tech{} generates stage-targeted test inputs using defect-guided mutation, web-search-assisted fidelity assurance, and diversity-aware seed maintenance.
Experiments across 16 framework--model--dataset settings show that \tech{} achieves 3.21 times the failure-discovery throughput of the strongest baseline under identical testing budgets.
Human and LLM-assisted evaluations demonstrate that most generated inputs preserve the original information need and necessary contextual constraints.
Moreover, model fine-tuning experiments show consistent-answer performance improvement, highlighting their value for robustness testing and model improvement.
\end{abstract}




\maketitle

\section{Introduction}
\label{sec:introduction}

Large language models (LLMs) have evolved into autonomous agents that plan, use external tools, and make multi-round decisions.
Among these applications, Deep Research (DR) systems have emerged as a promising paradigm for complex information-seeking tasks, integrating long-horizon planning, iterative web search, multi-source evidence collection, and information synthesis~\cite{huang2025deepresearchagentssystematic}.
Building on earlier web-assisted question-answering systems~\cite{DBLP:journals/corr/abs-2112-09332,DBLP:conf/naacl/ShaoJK0KL24,DBLP:conf/acl/QinCJYLZLHDWXQL23}, DR systems autonomously formulate subquestions, retrieve relevant evidence, and synthesize final answers.
However, their long and interactive execution processes introduce substantial reliability challenges.

Unlike conventional question-answering systems, DR systems involve multiple interdependent stages, including subquestion formulation, online search, and final reasoning.
Errors introduced at an early stage may propagate through subsequent stages and affect the final answer.
For example, incorrect entity identification may redirect subsequent searches, while minor variations in search expressions may lead to irrelevant evidence.
Such reliability challenges are also reflected in execution consistency:  WebSailor~\cite{li2025websailor} achieves a Pass@1 of 18.34\% on FRAMES-ZH~\cite{krishna2024frames}, but only 6\% of questions are answered correctly across three independent runs (Consistent@3).
Moreover, single-run accuracy on fixed questions may overlook robustness weaknesses, as a system that correctly answers an original question may fail under answer-preserving variations in user questions or intermediate search inputs.
We refer to such pass-to-fail behavioral changes as \textbf{robustness failures}.

Existing studies have primarily evaluated DR systems using benchmarks for complex information-seeking capabilities~\cite{krishna2024frames,wei2025browsecomp,du2025deepresearchbench,futuresearch2025deepresearchbench}.
For instance, DeepResearch Bench~\cite{du2025deepresearchbench} evaluates research report quality, evidence coverage, and citation accuracy.
However, these static benchmarks rely on predefined questions, which may become outdated and provide limited insights into robustness under controlled input variations.
To address these limitations, DR-Arena~\cite{gao2026drarena} dynamically constructs research tasks from live web information and adaptively increases task difficulty to explore system capability boundaries.
Recent studies have also investigated the robustness of DR systems against misleading or manipulated external web evidence~\cite{zhu2026misknowagent,dong2025safesearch,nie2026drnoise,pan2026forge,bi2026haegeo}.
Despite these advances, existing approaches primarily focus on overall task-solving capabilities or robustness against unreliable external information, leaving robustness weaknesses arising within DR execution processes largely unexplored.
Moreover, they lack systematic testing strategies informed by recurring defects across different stages of DR execution.
Such weaknesses are equally important, as even minor perturbations in user questions or intermediate search inputs may disrupt question interpretation, evidence retrieval, and final reasoning without manipulating external web content.
In practical applications such as scientific literature reviews and business analysis, these failures may cause critical evidence to be overlooked or incorrect conclusions to be produced.
When such outputs inform subsequent research or decision-making, seemingly minor input variations can lead to misleading analyses and potentially costly decisions.

To systematically investigate the robustness weaknesses of DR systems and guide targeted test generation, we conduct an empirical study of their execution trajectories.
Specifically, we collect 1,932 execution trajectories from three representative DR systems developed by leading industrial and academic research groups, i.e., DeepDiver~\cite{shi2025deepdiveradaptivesearchintensity}, WebSailor~\cite{li2025websailor}, and WebExplorer~\cite{liu2025webexplorer}.
We evaluate these systems on three benchmarks, i.e., Bamboogle~\cite{aksitov2023rest}, FRAMES-ZH~\cite{shi2025deepdiveradaptivesearchintensity}, and BrowseComp-ZH~\cite{zhou2025browsecompzh}, covering diverse information-seeking tasks ranging from compositional multi-hop reasoning to challenging open-web information retrieval.
Following statistical sampling~\cite{singh1996elements}, we manually analyze 384 trajectories to investigate two aspects: the distribution of observed failure origins across DR execution stages (\textbf{RQ1}) and the recurring defect characteristics underlying these failures (\textbf{RQ2}).
Our empirical analysis yields the following key findings.
For RQ1, subquestion formulation accounts for the largest proportion of classified failures (65.6\%), followed by online search (24.2\%) and final reasoning (10.2\%), indicating that early-stage errors are a major source of observed failures.
For RQ2, we identify nine recurring defect categories, revealing three major weaknesses: early interpretation errors can misdirect subsequent research, online search is sensitive to non-semantic variations and distracting evidence, and final reasoning may fail even when the required facts have been retrieved.

However, translating these empirical findings into effective robustness tests presents three technical challenges.
First, the interdependent execution stages of DR systems make it difficult to construct targeted perturbations that expose specific weaknesses without introducing unrelated changes.
Second, modifying natural-language questions or intermediate search inputs may inadvertently alter essential facts, contextual constraints, or reference answers, challenging test validity.
Third, the vast input space and costly multi-round execution of DR systems make it difficult to efficiently explore diverse robustness weaknesses while avoiding redundant test generation.
To address these challenges, we propose \tech{}, an empirically motivated, defect-guided fuzzing framework for robustness testing of DR systems.
Specifically, \tech{} first initializes a seed pool with questions the target DR system consistently answers correctly, along with their execution trajectories.
For each selected seed, a dual-stage mutation selection strategy determines whether to modify the original user question or an intermediate search input and then selects an applicable rule from nine defect-guided mutation rules derived from our empirical findings.
An LLM-based mutation operator generates test inputs via multi-round planning and online fact verification, aiming to preserve the original information need and reference answer.
The generated inputs are subsequently executed on the target DR system, and a semantic answer oracle identifies potential pass-to-fail robustness failures.
Finally, diversity-aware seed maintenance retains promising non-failing inputs for further mutation, enabling continued exploration of potential robustness failures.

We further evaluate \tech{} through three research questions examining its failure-detection effectiveness (\textbf{RQ3}), the contributions of its core components (\textbf{RQ4}), and its performance in human and LLM-assisted evaluations (\textbf{RQ5}).
We conduct extensive experiments across 16 framework--model--dataset settings, combining two DR frameworks (i.e., WebSailor~\cite{li2025websailor} and WebExplorer~\cite{liu2025webexplorer}) and their corresponding backbone models on four evaluation datasets (i.e., FRAMES~\cite{krishna2024frames}, FRAMES-ZH~\cite{shi2025deepdiveradaptivesearchintensity}, BrowseComp~\cite{wei2025browsecomp}, and OpenSeeker~\cite{du2026openseeker}).
Compared with DR-Arena and RobustAlpacaEval, \tech{} achieves 3.21 times the failure-discovery throughput of the strongest baseline under identical 24-hour testing budgets.
Ablation experiments confirm the effectiveness of its core components, while human and LLM-assisted evaluations demonstrate high mutation fidelity and assess the reliability of automated correctness judgments.
Beyond failure detection, fine-tuning WebSailor-32B with \tech{}-generated data improves three-run consistent correctness by up to 9.13 percentage points on held-out benchmarks, demonstrating their potential for improving DR systems.
Overall, these results highlight the effectiveness of empirically motivated, defect-guided fuzzing for robustness testing and reliability improvement of DR systems.

\smallskip
\noindent
\textbf{Contributions.}
In summary, this work makes the following contributions:
(1) We conduct an empirical study of 1,932 execution trajectories from three representative DR systems, identifying nine recurring defect categories across three execution stages.
(2) We propose \tech{}, an empirically motivated, defect-guided fuzzing framework that integrates defect-guided mutation, web-search-assisted fidelity assurance, and diversity-aware seed maintenance to systematically expose robustness failures.
(3) We conduct comprehensive automated and human evaluations, demonstrating superior failure-detection effectiveness over existing approaches and the potential of \tech{}-generated data to improve DR system performance through fine-tuning.
To support future research, we publicly release our replication package~\cite{our/url}, which includes all necessary code and resources to reproduce the results.

\section{Background and Related Work}
\label{sec:background}

This section reviews the architectures of DR systems and existing approaches to evaluating and testing their reliability.

\smallskip
\noindent
\textbf{Deep Research Systems.}
DR systems leverage LLMs to solve complex information-seeking tasks through long-horizon planning, iterative web search, and evidence synthesis~\cite{huang2025deepresearchagentssystematic}.
Based on their execution architectures, existing DR systems can be broadly categorized into ReAct-based and multi-agent-based systems.

ReAct-based DR systems typically employ a single agent to iteratively decompose questions, formulate search queries, retrieve web evidence, and update their reasoning following the ReAct paradigm~\cite{yao2023react}.
Recent studies have enhanced their research capabilities through different strategies.
For example, WebDancer~\cite{wu2025webdancer} combines supervised fine-tuning and reinforcement learning, while DeepDiver~\cite{shi2025deepdiveradaptivesearchintensity} introduces adaptive search intensity scaling.
WebThinker~\cite{li2025webthinker} integrates reasoning, web exploration, and report writing, whereas WebSailor~\cite{li2025websailor} and WebExplorer~\cite{liu2025webexplorer} enhance long-horizon research through uncertainty reduction and extended web exploration, respectively.
Despite these advances, sequential DR systems remain vulnerable to error propagation, where early decomposition or retrieval errors may affect subsequent decisions~\cite{wu2025resum}.
Multi-agent-based DR systems instead distribute research tasks among collaborating agents, typically involving a planner, specialized research agents, and a writer.
For example, MindSearch~\cite{chen2024mindsearch} dynamically constructs research graphs to coordinate parallel WebSearchers, while DeepResearcher~\cite{zheng2025deepresearcher} employs specialized agents for information acquisition and cross-source validation.
DeepDiver-V2~\cite{openpangu2025deepdiverv2} adopts a planner-centric architecture to coordinate information seekers and a writer, whereas MiroFlow~\cite{su2026miroflow} introduces flexible agent orchestration for complex research workflows.

Although multi-agent architectures improve task decomposition and information coverage, their individual research agents often still rely on ReAct-based execution, where errors may propagate through subsequent research steps.
In this study, we focus on three representative ReAct-based DR systems, i.e., DeepDiver, WebSailor, and WebExplorer.
These systems adopt different strategies for long-horizon information seeking and have been recognized or adopted in subsequent studies of web research agents~\cite{wu2025resum,zhang2026argus,pan2026contextpilot}.
Moreover, their available execution trajectories enable us to systematically investigate recurring failure patterns across subquestion formulation, online search, and final reasoning.

\smallskip
\noindent
\textbf{Deep Research System Evaluation.}
Existing DR evaluation approaches primarily rely on static benchmarks to assess information retrieval, multi-hop reasoning, and evidence synthesis.
Early benchmarks include WebArena~\cite{zhou2024webarena} for web interaction, Bamboogle~\cite{aksitov2023rest} for compositional question answering, FRAMES~\cite{krishna2024frames} for multi-step retrieval and reasoning, and BrowseComp~\cite{wei2025browsecomp} for challenging open-web information seeking.
More recent benchmarks evaluate complete research workflows.
For example, Du et al.~\cite{du2025deepresearchbench} introduce DeepResearch Bench with 100 expert-designed tasks to evaluate research quality and citation accuracy.
DeepResearch Arena~\cite{wan2025deepresearcharena} constructs tasks from academic seminars, while DeepWideSearch~\cite{lan2025deepwidesearch} evaluates reasoning depth and information coverage.
To improve evaluation reproducibility, FutureSearch~\cite{futuresearch2025deepresearchbench} provides a frozen search environment, while BrowseComp-Plus~\cite{chen2026browsecompplus} uses a fixed document corpus to disentangle retrieval effectiveness from agent capabilities.
However, static benchmarks rely on predefined questions and reference answers, which may become outdated as web information evolves or suffer data contamination.
To address these limitations, DR-Arena~\cite{gao2026drarena} dynamically generates research tasks from live web information and adaptively increases their difficulty to explore DR capability boundaries.

Beyond capability-oriented evaluation, recent studies investigate the robustness and security of DR systems under unreliable or adversarial web environments.
SafeSearch~\cite{dong2025safesearch} assesses search-agent safety under unreliable web content, while DRNOISE~\cite{nie2026drnoise} and MisKnow-Agent~\cite{zhu2026misknowagent} investigate the impact of misleading external evidence on research outcomes.
Similarly, FORGE~\cite{pan2026forge} examines how poisoned documents can redirect research planning, and HAE-GEO~\cite{bi2026haegeo} evaluates evidence verification and recovery under hierarchical web poisoning.
AgentFence~\cite{puppala2026agentfence} further investigates security vulnerabilities across different agent architectures.
Another line of research explores input-level robustness testing for QA systems and LLMs.
QATest~\cite{liu2022qatest} introduces multi-level semantic mutations for QA fuzzing, while QAQA~\cite{shen2022qaqa} employs sentence-level metamorphic relations to generate semantics-preserving test cases.
More recently, RobustAlpacaEval~\cite{cao2024worstprompt} generates semantically equivalent paraphrases to assess the robustness of LLMs under alternative input formulations.
These studies demonstrate the potential of systematic input mutation for evaluating model robustness.

Overall, existing studies primarily focus on task-solving capabilities, robustness against unreliable external information, or general-purpose input mutations, leaving robustness weaknesses arising within DR execution processes insufficiently explored.
Moreover, existing testing approaches do not explicitly leverage recurring defects across different execution stages to guide targeted test generation.
In contrast, \tech{} derives mutation strategies from empirically identified defects and systematically explores pass-to-fail robustness failures through defect-guided fuzzing.

\section{Empirical Study}
\label{sec:study}

To systematically characterize the failure origins and recurring defect patterns of DR systems, we conduct an empirical study of their execution trajectories.
Specifically, we investigate the following two research questions (RQs):
\begin{itemize}[leftmargin=*]
    \item \textbf{RQ1:} Where are the earliest unrecovered errors observed in failed DR execution trajectories?
    \item \textbf{RQ2:} What recurring defect characteristics underlie failures in DR systems?
\end{itemize}
The resulting findings provide empirical guidance for designing the defect-guided fuzzing framework presented in Section~\ref{sec:methodology}.

\subsection{Data Collection and Annotation}
\label{sec:study_design}

\begin{figure*}[t]
  \centering
  \includegraphics[width=1\linewidth]{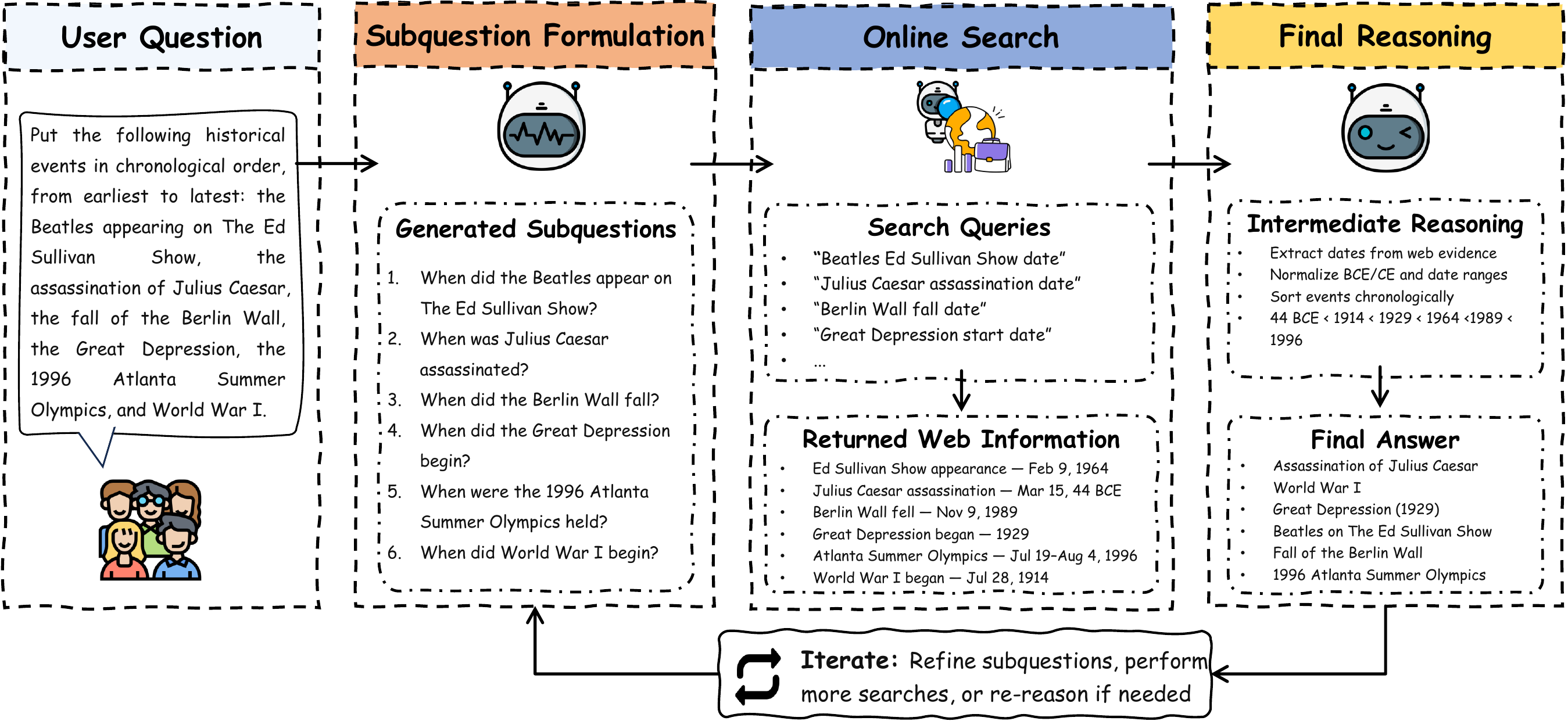}
  \caption{Illustrative execution trajectory of a Deep Research system.} 
  \label{fig:trajectory}
\end{figure*}

\textbf{Deep Research systems.}
To ensure the credibility and representativeness of our study, we select three representative ReAct-based DR systems: DeepDiver~\cite{shi2025deepdiveradaptivesearchintensity}, WebSailor~\cite{li2025websailor}, and WebExplorer~\cite{liu2025webexplorer}.
DeepDiver is a proprietary industrial system whose execution trajectories are provided by our industrial collaborators, while WebSailor and WebExplorer are open-source research frameworks.
As shown in Fig.~\ref{fig:trajectory}, these systems iteratively formulate subquestions, perform online searches, and synthesize retrieved evidence into final answers.
Their observable execution trajectories record intermediate reasoning, search queries, retrieved information, and final answers, enabling us to investigate observed failure origins rather than relying solely on final-answer correctness.

\smallskip
\noindent
\textbf{Benchmarks.}
We evaluate the selected DR systems on three benchmarks covering complementary information-seeking scenarios: Bamboogle~\cite{aksitov2023rest}, FRAMES-ZH~\cite{shi2025deepdiveradaptivesearchintensity}, and BrowseComp-ZH~\cite{zhou2025browsecompzh}.
Bamboogle consists of compositional questions that require systems to connect information from multiple sources.
FRAMES-ZH evaluates information retrieval and multi-step reasoning in Chinese information-seeking tasks, while BrowseComp-ZH contains difficult-to-find factual questions that typically require iterative browsing and evidence integration.
In total, we collect 1,932 execution trajectories.
We employ DeepSeek-V3.2~\cite{deepseekai2025deepseekv32} to assess answer correctness, requiring each answer to satisfy all constraints and requested components of the original question.
Following statistical sampling~\cite{singh1996elements}, we randomly select 384 incorrect trajectories for manual analysis, corresponding to a 95\% confidence level and a 5\% margin of error under standard sampling assumptions.

\smallskip
\noindent
\textbf{Failure-origin annotation.}
Following prior studies on trajectory-based failure attribution~\cite{pmlr-v267-zhang25cq,barke-etal-2026-agentrx}, we inspect the observable reasoning and tool-use trajectories to identify the earliest unrecovered errors associated with incorrect final answers.
Given a user question $x$ and a target DR system $\mathcal{F}(\cdot)$, we represent its execution trajectory as:
\begin{equation}
\tau(x) = \left\langle x,\ \mathbf{F},\ \mathbf{S},\ \mathbf{R},\ y \right\rangle,
\end{equation}
where $\mathbf{F}$, $\mathbf{S}$, and $\mathbf{R}$ denote subquestion formulation, online search, and final reasoning, respectively, and $y=\mathcal{F}(x)$ is the final answer.
We examine three critical stages:
\begin{itemize}[leftmargin=*, nosep]
    \item \textbf{Subquestion Formulation.}
    The DR interprets the user question, identifies the relevant entities, relationships, and constraints, and decomposes the original task into intermediate subquestions to guide subsequent information retrieval.

    \item \textbf{Online Search.}
    Based on the formulated subquestions, the DR constructs search queries, retrieves relevant web information, and examines the returned webpage content.
    It then extracts and summarizes useful evidence to support subsequent research and reasoning.

    \item \textbf{Final Reasoning.}
    The DR integrates the collected evidence across multiple subquestions and performs the reasoning or operations required by the original task, such as counting, arithmetic, and chronological ordering, to produce the final answer.
\end{itemize}
For each incorrect trajectory, we examine the sequence of reasoning and tool-use actions to identify the earliest observable error that remains uncorrected and is associated with the incorrect final answer.
Specifically, we compare the formulated subquestions with the original task requirements, inspect the retrieved evidence, and trace how erroneous interpretations or information are carried into subsequent search and reasoning steps.
Errors that are subsequently corrected and no longer affect the final answer are excluded from primary failure attribution.
When an earlier error propagates into later stages, we attribute the failure to the stage where the error first appears rather than counting its downstream consequences separately.
For example, if a subquestion misidentifies the target entity and subsequent searches consistently retrieve evidence about the incorrect entity, leading to a wrong final answer, the failure is attributed to subquestion formulation.
As our empirical study aims to characterize recurring failure patterns and their propagation across DR execution stages, we adopt trajectory-based attribution based on observable execution evidence rather than counterfactual interventions to establish causal necessity.


\smallskip
\noindent
\textbf{Taxonomy construction.}
Following prior studies~\cite{DBLP:journals/tse/RenDHWLJ25,nourry2023human,wang2025refactoring,revoredo2021practice}, we adopt a similar iterative manual annotation procedure to identify and categorize recurring defects in DR systems.
We define a defect pattern as a recurring type of erroneous execution behavior characterized by similar triggering conditions and error manifestations.
We use the three execution stages as an initial shared vocabulary, while specific defect categories are derived inductively rather than predefined.
Specifically, before formal annotation, two authors, who are proficient in English and Chinese and familiar with multi-step web QA, independently examine the same 100 incorrect trajectories to identify recurring defect patterns.
They compare their judgments, discuss inconsistent interpretations, and iteratively refine the taxonomy and annotation guidelines until reaching a consensus.
After establishing a shared understanding of the criteria and forming an initial taxonomy, they divide the remaining trajectories approximately evenly and annotate them separately using the established taxonomy.
When an observed failure cannot be explained by any existing category, the annotators jointly determine whether it represents a new recurring defect.
A new category is introduced only when its underlying error mechanism is clearly distinguishable and occurs across multiple trajectories.
The annotation process is supervised by two industry practitioners, who also review and validate the resulting taxonomy.
Ultimately, we identify nine recurring defect categories, organized according to the three critical stages of DR execution.
Table~\ref{tab:drs_defects} summarizes their definitions and distributions.

\begin{table*}[t]
\centering
\caption{Taxonomy of defects identified in Deep Research systems.}
\label{tab:drs_defects}
\small
\renewcommand{\arraystretch}{1.15}
\setlength{\tabcolsep}{2pt}
\renewcommand{\tabularxcolumn}[1]{m{#1}}

\begin{adjustbox}{width=\linewidth,center}
\begin{threeparttable}
\begin{tabularx}{1.05\linewidth}{
    >{\raggedright\arraybackslash}m{0.4\textwidth}
    >{\raggedright\arraybackslash}X
    >{\raggedleft\arraybackslash}m{0.06\textwidth}}
\toprule
\multicolumn{1}{c}{Category} &
\multicolumn{1}{c}{Description} &
\multicolumn{1}{c}{Ratio} \\

\specialrule{\lightrulewidth}{\aboverulesep}{0pt}
\rowcolor{gray!15}
\multicolumn{2}{@{}l}{
    \rule[-0.8ex]{0pt}{4.2ex}
    \textbf{\normalsize Subquestion Formulation}
} & \textbf{65.6\%} \\

\addlinespace[2pt]

\textbf{D1:} Misleading or Ambiguous Wording
& Ambiguous expressions lead to incorrect interpretations of entities, temporal relations, or references.
& 18.8\% \\

\addlinespace[1pt]

\textbf{D2:} Obscure or Niche Description
& Indirect or uncommon descriptions make it difficult to identify the intended entity.
& 10.9\% \\

\addlinespace[1pt]

\textbf{D3:} Missing Necessary Context
& Essential constraints are omitted in subquestion formulation, resulting in ambiguous information needs.
& 18.5\% \\

\addlinespace[1pt]

\textbf{D4:} Attention Allocation Bias
& Essential clues are overlooked despite being available in the original question.
& 17.4\% \\

\specialrule{\lightrulewidth}{\aboverulesep}{0pt}
\rowcolor{gray!15}
\multicolumn{2}{@{}l}{
    \rule[-0.8ex]{0pt}{4.2ex}
    \textbf{\normalsize Online Search}
} & \textbf{24.2\%} \\

\addlinespace[2pt]

\textbf{D5:} Related Popular Topic Distraction
& Popular neighboring topics or incomplete evidence divert retrieval from the intended target.
& 4.7\% \\

\addlinespace[1pt]

\textbf{D6:} Retrieval Strategy Sensitivity
& Irrelevant terms or colloquial query expressions redirect retrieval away from the intended evidence.
& 5.7\% \\

\addlinespace[1pt]

\textbf{D7:} Translation Issue
& Ambiguous translations or multilingual expressions distort entity identification and retrieval direction.
& 2.6\% \\

\addlinespace[1pt]

\textbf{D8:} Unreadable Webpage Content
& Relevant webpages are inaccessible or incompletely parsed, leaving critical evidence unavailable.
& 11.2\% \\

\specialrule{\lightrulewidth}{\aboverulesep}{0pt}
\rowcolor{gray!15}
\multicolumn{2}{@{}l}{
    \rule[-0.8ex]{0pt}{4.2ex}
    \textbf{\normalsize Final Reasoning}
} & \textbf{10.2\%} \\

\addlinespace[2pt]

\textbf{D9:} Complex Research with Simple Operation
& The system retrieves the required facts but fails on simple counting, ordering, or arithmetic operations.
& 10.2\% \\

\bottomrule
\end{tabularx}
\end{threeparttable}
\end{adjustbox}
\end{table*}

\subsection{RQ1: Distribution of Observed Failure Origins}
\label{sec:study_rq1}

To characterize the distribution of observed failure origins, we attribute each incorrect trajectory to the execution stage where its earliest unrecovered error is identified.
Among the 384 examined incorrect trajectories, 252 (65.6\%) failures are attributed to subquestion formulation, followed by 93 (24.2\%) from online search and 39 (10.2\%) from final reasoning.
Subquestion formulation accounts for nearly two-thirds of the classified failures, indicating that early-stage question interpretation is a major source of errors in DR systems.
When a system incorrectly identifies an entity, relation, or task constraint, the resulting subquestions may direct subsequent searches toward irrelevant information.
Such errors can propagate through multiple research steps, making it difficult for the system to recover from an incorrect initial interpretation.
Online search represents the second-largest source of failures, suggesting that correctly understanding the information need does not necessarily guarantee successful evidence retrieval.
Errors in search query formulation or evidence acquisition may still prevent the system from obtaining the required information.
Although final reasoning accounts for a smaller proportion, its failures demonstrate that retrieving relevant evidence alone does not guarantee a correct answer, particularly when additional reasoning or calculations are required.
Overall, these results highlight the importance of examining robustness weaknesses across different stages of DR execution rather than focusing solely on initial question interpretation.
Notably, the reported proportions characterize the distribution of identified failure origins, rather than the failure probabilities of individual execution stages.
\finding{\textbf{Answering RQ1.}
Subquestion formulation is the most frequent origin of the classified failures (65.6\%), followed by online search (24.2\%) and final reasoning (10.2\%).
These findings highlight the importance of robustness testing across multiple execution stages, particularly the early stages where errors may propagate through subsequent research steps.}

\subsection{RQ2: Recurring Defect Characteristics}
\label{sec:study_rq2}

To characterize the underlying weaknesses of DR systems, we further analyze the recurring defect patterns in the incorrect execution trajectories.
Our analysis identifies nine recurring defect categories (D1--D9), comprising four in subquestion formulation, four in online search, and one in final reasoning, as summarized in Table~\ref{tab:drs_defects}.
The relative frequencies of these defects may vary with backbone LLMs, execution strategies, and retrieval configurations.
As our study does not independently vary these factors, the reported distributions characterize the evaluated DR configurations rather than universal defect frequencies.
In the following, we characterize each category in terms of its triggering conditions, observed failure behaviors, and distinguishing characteristics, illustrated with representative cases.

\smallskip
\noindent
\textbf{Subquestion Formulation} \textit{(65.6\%)}.
This stage involves interpreting the entities, relationships, and constraints in a user question to determine intermediate information needs.
Defects at this stage may misdirect subsequent research before relevant evidence is retrieved.
We identify four recurring categories, as follows:

\textit{{D1: Misleading or Ambiguous Wording}} \textit{(18.8\%)}.
Questions containing ambiguous temporal, referential, or relational expressions may cause DR systems to adopt incorrect interpretations despite sufficient contextual clues.
For example, when asked about the number of female athletes participating in an Olympics associated with the main cauldron of the 1984 Los Angeles Games, a DR system incorrectly assumed that the target event occurred after 1984, overlooking the intended reference to an earlier event in 1932.
This defect manifests as premature interpretation of ambiguous expressions, steering subsequent research toward incorrect entities or temporal relationships.

\textit{{D2: Obscure or Niche Description}} \textit{(10.9\%)}.
Some questions describe target entities through indirect or uncommon clues rather than explicit names.
Although the complete question may contain sufficient clues to identify the target, DR systems can struggle to resolve these descriptions into precise entities.
For example, describing a target as ``a lake in California that expanded around 1966'' may lead the system to formulate broad search queries without identifying the intended lake.
Unlike ambiguous wording, this defect primarily arises from difficulties in recovering an entity from indirect descriptions, which can misdirect subsequent information retrieval.

\textit{{D3: Missing Necessary Context}} \textit{(18.5\%)}.
DR systems may omit essential qualifiers or contextual constraints when decomposing the original question into intermediate information needs.
Such omissions can make generated subquestions ambiguous even when the original task provides sufficient information.
For example, when searching a film’s in-story setting, the system may simplify ``American Pie 4 story setting time'' to ``American Pie 4 story time'', omitting the qualifier ``setting'' and potentially retrieving film's release date instead of its narrative period.
Unlike D1, where existing clues are misinterpreted, this defect arises when necessary information is lost during subquestion formulation, causing subsequent searches to pursue underspecified or incorrect targets.

\textit{{D4: Attention Allocation Bias}} \textit{(17.4\%)}.
Even when all necessary clues are present, DR systems may focus on salient but less relevant information while overlooking essential relationships.
For example, one question connects a famous portrait, a river mentioned in a Chinese song, and a legislative meeting in the lyricist's ancestral hometown.
Answering it requires identifying the river to establish the intended relationship, but the system concentrates on the explicitly mentioned legislative meeting without formulating subquestions about the river.
Unlike D3,  necessary information is not omitted from the task representation but receives insufficient attention during decomposition.

\smallskip
\noindent
\textbf{Online Search} \textit{(24.2\%)}.
Online search involves constructing search queries from intermediate information needs, retrieving web information, and extracting supporting evidence.
Even when the intended information need is correctly identified, failures may arise from query formulation, evidence selection, or webpage accessibility.
We identify four recurring categories, as follows:

\textit{{D5: Related Popular Topic Distraction}} \textit{(4.7\%)}.
Search results may favor popular topics that share keywords with the intended target, causing DR systems to prioritize frequently encountered but incomplete or irrelevant evidence.
For example, when investigating the number of Brendan Fraser's children, a system repeatedly focused on search results discussing two of his sons while overlooking evidence about his third child.
Subsequent searches reinforced this incomplete interpretation rather than verifying the full set of relevant information.
This defect reflects the influence of topic popularity on evidence selection and iterative search direction.

\textit{{D6: Retrieval Strategy Sensitivity}} \textit{(5.7\%)}.
The effectiveness of online search can be sensitive to query wording and structure, even when the underlying information need remains unchanged.
We observe two manifestations.
First, unrelated or overly prominent terms may dilute critical search keywords.
For example, in a question concerning a \emph{Suicide Squad} spin-off, the system emphasized terms associated with the original film and its filming location while overlooking the spin-off itself, resulting in irrelevant search results.
Second, colloquial expressions may be carried directly into search queries instead of being reformulated into precise retrieval terms.
For instance, an expression such as ``Where did they shoot the Peacemaker show?'' may lead to less focused retrieval than ``Peacemaker TV series filming location''.
Both manifestations demonstrate query formulation may redirect retrieval despite preserving the underlying information need.

\textit{{D7: Translation Issue}} \textit{(2.6\%)}.
Multilingual search may fail when translated expressions do not preserve the canonical names or intended meanings of target entities.
For example, translating \emph{Postcolonial Love Poem} into a generic Chinese expression equivalent to ``postcolonial love poetry'' may cause search results to focus on a broad literary topic rather than the specific publication.
In another case, an English reformulation of a geographical query led the system to identify Belfast Harbour instead of the River Lagan.
This defect arises from translation ambiguity, distorting entity identification and redirecting subsequent retrieval.

\textit{{D8: Unreadable Webpage Content}} \textit{(11.2\%)}.
Even when a DR system locates relevant webpages, it may fail to access or extract the information needed to answer the question.
Such failures can arise from dynamic rendering, anti-crawling mechanisms, inaccessible URLs, or limitations of browsing tools.
For example, a browsing tool may retrieve a relevant webpage but omit the passage containing the required fact, leaving the system with insufficient evidence for subsequent reasoning.
Unlike the preceding search-related defects, this category primarily concerns the accessibility and extraction of web evidence rather than the relevance of search queries or retrieved results.

\smallskip
\noindent
\textbf{Final Reasoning} \textit{(10.2\%)}.
Final reasoning integrates evidence collected throughout the research process and performs the operations required to produce the final answer.
We identify one recurring defect category at this stage, as follows:

\textit{{D9: Complex Research with Simple Operation}} \textit{(10.2\%)}.
DR systems may successfully retrieve the required facts but fail to perform relatively simple operations after a long research process.
For example, when asked to sum the numerical values of the years marking the beginning and end of John Lennon's life, a system may incorrectly sum the individual digits of these years rather than the years themselves.
Similar failures occur in tasks involving counting or chronological ordering, even when the necessary evidence has already been collected.
These failures suggest that long research trajectories can interfere with the preservation of task requirements, even when the final operation itself is straightforward.
\finding{
\textbf{Answering RQ2.}
Our analysis identifies nine recurring defect categories across three critical stages of DR execution.
These defects reveal distinct weaknesses in information-need interpretation, evidence retrieval, and final reasoning, providing empirical guidance for designing targeted robustness tests rather than relying solely on general-purpose input perturbations.
}

\subsection{Implications for Defect-Guided Fuzzing}
\label{sec:study_implications}

The findings from RQ1 and RQ2 reveal three recurring weaknesses across critical stages of DR execution, providing insights into the design of targeted robustness tests.

\textbf{L1: Early Interpretation Errors Can Propagate Across Research Stages.}
Our stage-level analysis reveals that subquestion formulation accounts for the largest proportion of classified failures.
Once a DR system misinterprets an entity, relation, or task constraint, subsequent searches may reinforce the incorrect interpretation rather than correct it.
This finding motivates targeted mutations of user questions to examine whether DR systems can consistently recover the intended information need from alternative expressions.

\textbf{L2: Online Search is Sensitive to Query Formulation and Retrieved Evidence.}
Our defect analysis shows that irrelevant terms, colloquial expressions, multilingual ambiguity, and distracting neighboring topics can redirect retrieval even when the underlying information need remains unchanged.
This finding motivates targeted mutations of intermediate search inputs to examine how variations in query formulation affect evidence retrieval and subsequent reasoning.
In contrast, failures caused by inaccessible webpages primarily concern the external environment and require different testing interventions.

\textbf{L3: Correct Evidence Retrieval Does Not Guarantee Accurate Final Reasoning.}
Our analysis identifies cases where DR systems successfully retrieve the required facts but fail to perform simple counting, ordering, or arithmetic operations after lengthy research processes.
This finding motivates targeted tests that introduce simple reasoning demands into complex research tasks, while ensuring that the intended output and reference answer remain consistent.

\smallskip
\noindent
\textbf{Implications for Robustness Testing.}
These findings suggest three key requirements for effective DR robustness testing.
First, the identified failure stages and defect characteristics provide empirically grounded, yet non-exhaustive, guidance for targeting observed robustness weaknesses.
Second, generated test inputs should preserve the original information need and reference answer, requiring careful verification of factual and semantic constraints.
Third, testing should encourage diversity across seed questions and mutation strategies to explore different potential failure behaviors rather than repeatedly exposing similar weaknesses.
Together, these requirements motivate systematic, defect-guided robustness testing, which complements broader evaluations by targeting empirically observed weaknesses without claiming exhaustive coverage of all possible failures.

\section{Approach}
\label{sec:methodology}

\begin{figure*}[t]
  \centering
  \includegraphics[width=1\linewidth]{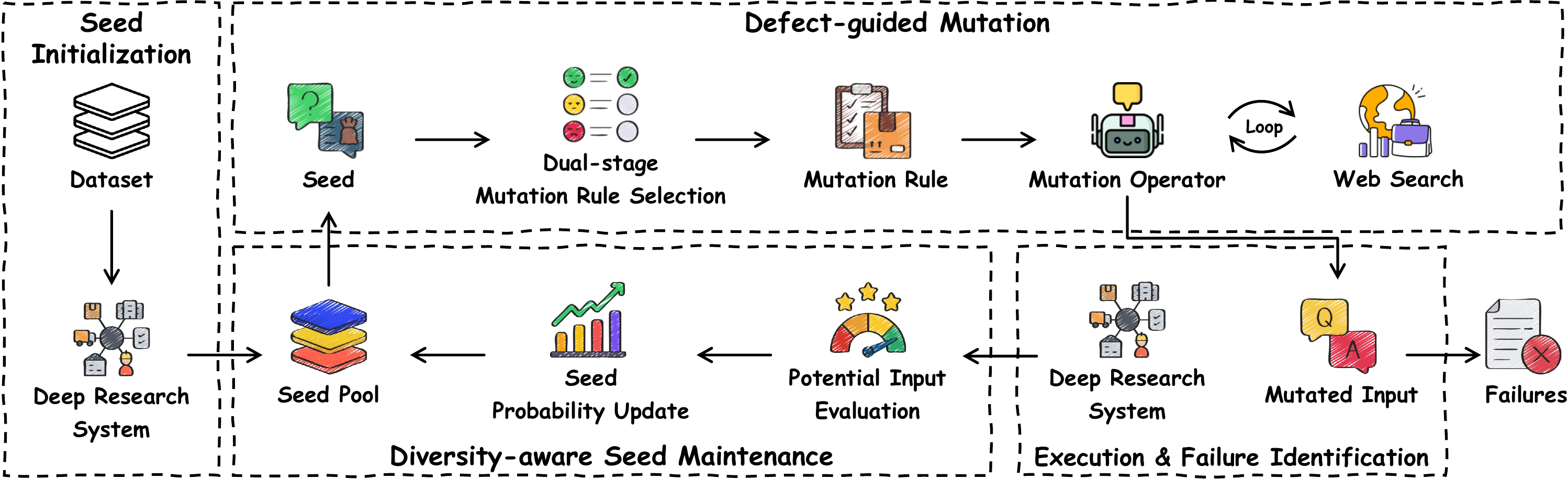}
  \caption{The overview of \tech{}} 
  \label{fig:overview}
\end{figure*}

This section presents \tech{}, a defect-guided fuzzing framework for robustness testing of Deep Research systems.
As illustrated in Fig.~\ref{fig:overview}, \tech{} initializes a seed pool with correctly answered questions and iteratively generates stage-targeted test inputs based on empirically identified defect characteristics.
The generated inputs are executed on the target DR system to identify potential robustness failures, while promising non-failing inputs are retained for further mutation to support continued exploration.
In the following, we first introduce the diversity measurement and web-search-assisted fidelity assurance mechanism, followed by the defect-guided fuzzing process, including seed initialization, mutation, execution, and failure identification, and seed maintenance.

\subsection{Diversity Measurement}
\label{sec:diversity}
Effective robustness testing should not only expose numerous failures but also explore diverse underlying weaknesses.
However, repeatedly mutating similar inputs may lead to redundant failures, while directly determining whether different failures originate from distinct defects is challenging due to the multi-stage execution and open-ended outputs of DR systems.
To address this challenge, inspired by prior studies that approximate fault diversity using observable information~\cite{DBLP:conf/icse/YouWCLL23,DBLP:conf/issta/XieMJXCLZLYS19}, \tech{} employs a lightweight diversity measurement based on two complementary aspects: static diversity and dynamic diversity.

\textbf{Static Diversity.}
The first aspect considers static initial seeds from which failure-inducing inputs are generated.
The intuition is that different initial questions typically involve distinct information needs, entity relationships, search paths, and reasoning requirements, making derived test inputs more likely to expose different robustness weaknesses~\cite{DBLP:journals/tosem/0003WWYZY20,DBLP:conf/sigsoft/WangYCLZ20,DBLP:conf/issta/XieMJXCLZLYS19,liu2025cq2a,shen2022qaqa}.
Specifically, given two failure-inducing inputs $x_i^a$ and $x_j^b$ generated through a series of mutations from different initial seeds $x_0^a$ and $x_0^b$, respectively, we consider them more likely to expose different weaknesses, even when they are generated using the same mutation rule.
Here, initial seeds refer to inputs admitted into the seed pool before fuzzing begins; formally, $\mathit{sid}(x_0)$ denotes the identifier of initial seed $x_0$.

\textbf{Dynamic Diversity.}
The second aspect considers dynamic mutation-induced failure behaviors, complementing static information from initial seeds.
Unlike classification systems, where failure behaviors can be characterized by prediction transitions~\cite{DBLP:conf/icse/YouWCLL23,DBLP:conf/issta/XieMJXCLZLYS19}, DR systems produce open-ended answers, making such transitions hard to define.
To approximate failure behaviors here, \tech{} uses the applied mutation rule as an observable proxy.
The intuition is that different mutation rules target distinct defect characteristics and may therefore expose different robustness weaknesses, even when applied to the same initial seed.
For example, obscure-description mutation examines entity identification under indirect descriptions, whereas translation-sensitive mutation examines whether multilingual expressions affect evidence retrieval.
Accordingly, inputs generated using different mutation rules are considered more likely to exercise distinct failure mechanisms.
Formally, for a generated input $x_i$, we denote the mutation rule applied in its latest mutation step as $\mu(x_i)$.

Combining these two aspects, \tech{} represents the diversity state of a generated input as:
\begin{equation}
D(x_i)=\langle sid(x_0),\mu(x_i)\rangle,
\end{equation}
This representation approximates potential failure diversity through the originating information need and mutation strategy, without requiring precise identification of underlying defects.
As described in Section~\ref{sec:seed_maintenance}, \tech{} uses these diversity states to guide the retention of promising non-failing inputs and reduce redundant exploration.

\subsection{Web Search-Assisted Fidelity Assurance}
\label{sec:fidelity}

Another challenge in DR system fuzzing is how to preserve the fidelity of generated test inputs.
Unlike conventional structured inputs, natural-language questions contain interconnected entities, relations, and contextual constraints, making modification prone to unintended semantic changes.
For example, replacing an explicit entity name with an indirect description may introduce multiple possible targets, while adding background information or translating entity names may alter the intended reference.
Such mutations can invalidate robustness testing if they change the original information need or make the reference answer ambiguous.

To address this challenge, \tech{} employs a web-search-assisted LLM mutation operator to generate test inputs while preserving essential factual and semantic constraints.
The operator is guided by a structured prompt specifying the assigned mutation rule, preservation constraints, and positive and negative examples.
These constraints require generated inputs to retain answer-determining information, preserve the intended target and reference answer, and avoid fabricated facts or unintended answer changes.
Specifically, the operator follows a multi-round procedure consisting of four steps:
\begin{itemize}[leftmargin=*]
    \item \textbf{\textit{Step 1: Mutable Subject Identification.}}
    The operator first identifies entities, relations, temporal expressions, attributes, or search terms that can be modified according to the assigned mutation rule.
    Meanwhile, it distinguishes these mutable elements from essential constraints that must be preserved to maintain the original information need.
    For example, in a historical-event ordering question, event names may be replaced with equivalent descriptions, but the target events and their ordering requirement must remain unchanged.

    \item \textbf{\textit{Step 2: Mutation Direction and Constraint Determination.}}
    Based on the identified elements, the operator explores potential mutation directions and determines the constraints that each modification must satisfy.
    For example, obscure-description mutation may replace an explicit entity name with an indirect description, whereas attention-distraction mutation may introduce additional but irrelevant factual information.
    The operator selects a suitable mutation direction while maintaining the original task requirements.

    \item \textbf{\textit{Step 3: Online Fact Verification.}}
    Before generating the mutated input, the operator invokes a web-search tool to verify the factual validity of the proposed modifications.
    Retrieved evidence is used to check entity descriptions, contextual relationships, temporal constraints, and multilingual expressions, with additional searches performed when necessary.
    This step provides external evidence for mutations involving obscure or specialized knowledge that may be difficult to verify using the LLM's parametric knowledge alone.

    \item \textbf{\textit{Step 4: Generation and Inspection.}}
    Using the verified information, the operator generates the mutated input and inspects its consistency with the original information need and mutation constraints.
    Depending on the mutation target, it produces either a modified user question or an intermediate search input while preserving the preceding execution context.
    The resulting mutation and supporting evidence are recorded for subsequent testing and inspection.
\end{itemize}
Through constraint-guided mutation and online fact verification, \tech{} aims to reduce factual inaccuracies and unintended semantic changes in generated test inputs.
By maintaining the original information need and reference answer, this mechanism improves the validity of pass-to-fail robustness testing, although complete preservation cannot be guaranteed.

\subsection{Defect-Guided Fuzzing Process}
\label{sec:fuzzing_process}

The large input space of DR systems makes unguided mutation inefficient, as many generated inputs either fail to expose new weaknesses or inadvertently alter the original information need.
To improve failure discovery, \tech{} adopts a defect-guided fuzzing process based on the empirical findings in Section~\ref{sec:study}.
Following the standard fuzzing workflow~\cite{DBLP:conf/icse/YouWCLL23,DBLP:conf/issta/XieMJXCLZLYS19}, the process consists of four steps: seed initialization, mutation, execution and failure identification, and seed maintenance.

\subsubsection{Seed Initialization}
\label{sec:seed_initialization}
To discover robustness weaknesses through answer-preserving input mutations, \tech{} initializes its seed pool with questions that the target DR system can answer correctly.
These questions establish a passing baseline, allowing failures observed on mutated inputs to be distinguished from errors already present in the original questions.
To reduce the influence of execution randomness, \tech{} retains only questions that the target DR system answers correctly across three independent executions.
For each retained question $x$, \tech{} constructs a seed $s$ consisting of the question, its reference answer $a$, and a corresponding execution trajectory $\tau(x)$:
\begin{equation}
s=\langle x,a,\tau(x)\rangle.
\end{equation}
The trajectory records the system's reasoning, generated subquestions, search actions, retrieved evidence, and final answer, as defined in Section~\ref{sec:study_design}.
By retaining both the original question and its execution trajectory, \tech{} enables subsequent mutations of either the user question or intermediate search inputs, providing the basis for stage-targeted robustness testing.

\begin{table*}[t]
\centering
\caption{Defect-guided mutation rules in \tech{}.}
\label{tab:mutation_rule}
\small
\renewcommand{\arraystretch}{1.15}
\setlength{\tabcolsep}{2pt}
\renewcommand{\tabularxcolumn}[1]{m{#1}}

\begin{adjustbox}{width=\linewidth,center}
\begin{threeparttable}
\begin{tabularx}{1.05\linewidth}{
    >{\raggedright\arraybackslash}m{0.25\textwidth}
    >{\raggedright\arraybackslash}X}
\toprule
\multicolumn{1}{c}{{\small Mutation Rule}} &
\multicolumn{1}{c}{{\small Description}} \\

\specialrule{\lightrulewidth}{\aboverulesep}{0pt}

\rowcolor{gray!15}
\multicolumn{2}{@{}l}{
    \rule[-0.8ex]{0pt}{4.2ex}
    \textbf{\normalsize Subquestion Formulation}
} \\
\addlinespace[2pt]

\textbf{M1:} Misleading-ambiguity
& Rephrases temporal, referential, or relational expressions into potentially misleading forms while preserving their original meaning, testing the system's ability to resolve ambiguity. \\

\addlinespace[1pt]

\textbf{M2:} Obscure-description
& Replaces explicit entity or event names with verified indirect descriptions that uniquely identify the same target, testing entity resolution during subquestion formulation. \\

\addlinespace[1pt]

\textbf{M3:} Missing-context
& Makes contextual qualifiers less explicit while retaining answer-determining information, testing whether the system preserves essential constraints during subquestion formulation. \\

\addlinespace[1pt]

\textbf{M4:} Attention-distraction
& Introduces truthful but irrelevant background information to reduce the salience of critical clues, testing whether the system maintains attention to essential information. \\

\specialrule{\lightrulewidth}{\aboverulesep}{0pt}

\rowcolor{gray!15}
\multicolumn{2}{@{}l}{
    \rule[-0.8ex]{0pt}{4.2ex}
    \textbf{\normalsize Online Search}
} \\
\addlinespace[2pt]

\textbf{M5:} Related-hot-topic
& Introduces popular neighboring topics into search expressions to test whether related but irrelevant information distracts retrieval from the intended target. \\

\addlinespace[1pt]

\textbf{M6:} Unrelated-word
& Injects irrelevant or high-frequency terms into search expressions to test whether noisy keywords redirect retrieval away from the intended evidence. \\

\addlinespace[1pt]

\textbf{M7:} Spoken-language
& Reformulates keyword-based search queries into colloquial expressions while preserving their meaning, testing sensitivity to alternative query formulations. \\

\addlinespace[1pt]

\textbf{M8:} Translation-sensitive
& Introduces alternative translations, mixed-language expressions, or non-standard entity names to test whether multilingual retrieval preserves the intended entities and relationships. \\

\specialrule{\lightrulewidth}{\aboverulesep}{0pt}

\rowcolor{gray!15}
\multicolumn{2}{@{}l}{
    \rule[-0.8ex]{0pt}{4.2ex}
    \textbf{\normalsize Final Reasoning}
} \\
\addlinespace[2pt]

\textbf{M9:} Complex-simple-combination
& Incorporates auxiliary counting, ordering, or arithmetic steps into complex research tasks, while explicitly preserving the original requested output and reference answer, testing the system's ability to perform simple operations after extensive research. \\

\bottomrule
\end{tabularx}
\end{threeparttable}
\end{adjustbox}
\end{table*}

\subsubsection{Defect-guided Mutation}
\label{sec:mutation}

Mutation is the core operation of \tech{}, transforming correctly answered seeds into new test inputs that target empirically identified robustness weaknesses.
Unlike general-purpose textual perturbations, \tech{} derives nine mutation rules from the defect characteristics identified in Section~\ref{sec:study}.
As summarized in Table~\ref{tab:mutation_rule}, these rules target different stages of DR execution.

\textbf{Defect-guided Mutation Rules.}
The nine mutation rules (M1--M9) are designed to expose weaknesses across subquestion formulation, online search, and final reasoning.
For subquestion formulation, M1--M4 are derived from D1--D4 and modify the original user question to examine whether the DR system correctly interprets entities, relationships, and contextual constraints.
For online search, M5--M8 are derived from D5--D7 and modify intermediate search-oriented inputs to investigate sensitivity to distracting topics, irrelevant terms, colloquial expressions, and multilingual ambiguity.
Notably, D6 (Retrieval Strategy Sensitivity) motivates two complementary rules, M6 (Unrelated-word mutation) and M7 (Spoken-language mutation), which target different aspects of query formulation.
For final reasoning, M9 (Complex-simple-combination mutation), derived from D9, introduces auxiliary counting, ordering, or arithmetic operations into complex research tasks while aiming to preserve the original information need and expected answer.
We do not directly design a mutation rule for D8 (Unreadable Webpage Content), as it primarily concerns external webpage accessibility and content extraction rather than weaknesses behind the reasoning process of the DR system itself.
Such environment-level failures are beyond the scope of our input-level fuzzing approach~\cite{futuresearch2025deepresearchbench,wang2026agentnoisebench,yin2026breakingweb}.
The mutation target depends on the execution stage being tested.
Rules targeting subquestion formulation and final reasoning operate on the original user question, requiring the DR system to restart its research process.
In contrast, online-search rules modify intermediate subquestions or search queries while preserving the preceding execution context.
These mutations are generated by the web-search-assisted operator described in Section~\ref{sec:fidelity}, which aims to maintain factual and semantic consistency with the original task.

\textbf{Dual-stage Mutation Rule Selection.}
Different mutation rules target distinct robustness weaknesses, making their selection important for effective failure discovery.
However, directly sampling from all available rules couples mutation opportunities to the number of rules associated with each target, potentially leading to uneven exploration.
To address this limitation, \tech{} adopts a dual-stage selection strategy that separates mutation target selection from mutation rule selection.
This design enables explicit control over mutation opportunities at different input levels while ensuring that each selected rule is compatible with its target.

Specifically, \tech{} considers two mutation target categories within a seed trajectory: the original user question $\mathtt{QUESTION}$ and intermediate search-oriented inputs $\mathtt{SUBQUESTION}$.
Given a selected seed $s$, let $\mathcal{G}(s)=\{\mathtt{QUESTION},\mathtt{SUBQUESTION}\}$ denote its available target categories.
First, \tech{} samples a target category $t\in\mathcal{G}(s)$ and then selects a concrete mutation point $p$ from:
\begin{equation}
\mathcal{P}_t(s)=
\begin{cases}
\{x\}, & t=\mathtt{QUESTION},\\
\{u_i\mid 1\leq i\leq m\}, & t=\mathtt{SUBQUESTION},
\end{cases}
\label{eq:mutation_target}
\end{equation}
where $x$ denotes the original user question and $u_i$ denotes the $i$-th intermediate search-oriented subquestion in the seed trajectory.
The target-selection probabilities are configurable to control the testing budget allocated to different input levels.
In our experiments, the two target categories are assigned equal selection probabilities.

Once the mutation target is determined, \tech{} selects an applicable mutation rule from the rule set associated with the selected target.
Let $\mathcal{R}(t)$ denote the rules applicable to target category $t$.
Specifically, $\mathcal{R}(\mathtt{QUESTION})$ contains M1--M4 and M9, while $\mathcal{R}(\mathtt{SUBQUESTION})$ contains M5--M8.
A mutation rule $r\in\mathcal{R}(t)$ is then randomly selected and applied to the chosen mutation point through the web-search-assisted mutation operator.
By separating target selection from rule selection, \tech{} avoids directly coupling the testing budget to the number of available mutation rules and supports balanced exploration of question-level and intermediate-search mutations.
Although we adopt a simple randomized selection strategy, our dual-stage design can be extended with adaptive optimization techniques, as explored in prior mutation scheduling studies~\cite{lyu2019mopt,jauernig2023darwin}, to dynamically adjust rule-selection probabilities based on failure-discovery feedback in future work.

\subsubsection{Execution and Failure Identification}
\label{sec:failure_identification}
After mutation, \tech{} executes the generated test input on the target DR system to determine whether it induces a robustness failure.
The execution strategy depends on the selected mutation target.
For question-level mutations, \tech{} replaces the original user question and restarts the DR system to produce a complete execution trajectory.
For intermediate-search mutations, it preserves the reasoning and tool-use history preceding the mutation point, replaces the selected subquestion or search query, and resumes the remaining execution.
This prefix-preserving execution strategy enables targeted testing of intermediate research stages without modifying the underlying model parameters.

After execution, \tech{} obtains the final answer $y'$ and employs a semantic answer oracle $\mathcal{O}$ to assess its correctness against the reference answer $a$.
Since DR systems produce open-ended responses containing explanations, citations, and alternative expressions, exact string matching is insufficient.
Instead, the oracle identifies the final conclusion and evaluates whether it satisfies all required answer components, allowing semantically equivalent expressions and superficial formatting differences.
Answers containing incorrect conclusions, missing required components, or no determinate final answer are classified as incorrect.
Given the original system output $y$ and the mutated output $y'$, a test input is identified as failure-inducing when:
\begin{equation}
\mathcal{O}(y,a)=1
\land
\mathcal{O}(y',a)=0,
\label{eq:failure_identification}
\end{equation}
where $\mathcal{O}(\cdot,a)\in\{0,1\}$ indicates whether the evaluated answer is correct.
This condition captures pass-to-fail transitions in which the original question is answered correctly, but the mutated input produces an incorrect answer.
For each identified failure, \tech{} records the originating seed, mutation target, applied rule, mutation history, generated input, execution trajectory, and oracle judgment to support subsequent analysis.

\subsubsection{Diversity-aware Seed Maintenance}
\label{sec:seed_maintenance}

Seed maintenance determines which correctly answered mutated inputs should be retained for further exploration.
Retaining every generated input may lead to redundant mutations, whereas discarding all non-failing inputs prevents the exploration of potentially informative mutation chains.
To balance these concerns, \tech{} employs two complementary mechanisms: potential input evaluation and seed probability update.
Both mechanisms leverage the diversity measurement introduced in Section~\ref{sec:diversity} to guide iterative fuzzing.

\textbf{Potential Input Evaluation.}
A straightforward retention strategy is to preserve inputs that introduce previously unseen diversity states.
However, the number of such states is limited, and state novelty alone cannot distinguish promising candidates once the same states are revisited.
To support continued exploration, \tech{} further evaluates the potential of non-failing inputs using their execution trajectories.
In our empirical study, incorrectly answered executions exhibit an average trajectory length of 19.86 steps, approximately 1.31 times that of correctly answered executions, which average 15.19 steps.
This observation suggests an association between longer research processes and execution failures, motivating the use of trajectory length as a lightweight heuristic for exploration potential.
Specifically, longer trajectories may involve more extensive subquestion formulation, web searching, and reasoning, providing richer intermediate contexts for subsequent mutations.
Accordingly, within each diversity state, a non-failing input is retained only when its trajectory length exceeds the historical maximum observed for that state.
Formally, given a generated input $x_i$, its execution trajectory $\tau(x_i)$, and diversity state $D(x_i)$, the seed pool $\mathcal{C}$ is updated as follows:

\begin{equation}
\mathcal{C}\leftarrow
\begin{cases}
\mathcal{C}\cup\{x_i\},
& \mathrm{Len}(\tau(x_i))>
\mathrm{Len}_{\max}(D(x_i)),\\
\mathcal{C},
& \text{otherwise},
\end{cases}
\end{equation}
where $\mathrm{Len}(\tau(x_i))$ denotes the length of the execution trajectory, measured by the total number of reasoning steps and tool calls, and $\mathrm{Len}_{\max}(D(x_i))$ denotes the historical maximum length among previously evaluated inputs in the same diversity state.
This strategy retains inputs that extend the explored trajectory-length frontier within their respective diversity states, reducing redundant seed accumulation while preserving candidates for higher-order mutation.

\textbf{Seed Probability Update.}
As the seed pool expands, \tech{} must allocate mutation opportunities among initial and newly retained seeds.
Repeatedly selecting the same seed may limit exploration diversity, while excessive mutation depth increases the risk of semantic drift.
To balance these factors, \tech{} employs an exponential decay strategy that adjusts seed selection weights by mutation depth and selection frequency.
Specifically, for each seed $s$, let $N_g(s)$ denote its mutation generation, i.e., the number of successive mutations from its initial seed, and $N_s(s)$ denote the number of times selected for mutation.
The unnormalized seed-selection weight is calculated as:
\begin{equation}
P(s) = P_{\mathrm{init}}
e^{-\alpha(w_1N_g(s)+w_2N_s(s))},
\label{eq:decay}
\end{equation}
where $P_{\mathrm{init}}$ and $P_{\mathrm{finish}}$
denote the initial and reference final selection weights,
respectively, and
$\alpha = \frac{1}{\mathrm{max\_times}}
\ln\left(\frac{P_{\mathrm{init}}}{P_{\mathrm{finish}}}\right)$
controls the decay rate.
The parameter $\mathrm{max\_times}$ controls the decay scale, specifying the weighted mutation activity at which the selection weight decreases from $P_{\mathrm{init}}$ to $P_{\mathrm{finish}}$.
This mechanism reduces the priority of deeply mutated or frequently selected seeds, limiting redundant exploration and mitigating the risk of substantial semantic deviations from the initial questions.
Meanwhile, $w_1$ and $w_2$ control the relative influence of mutation depth and selection frequency on the decay rate.
During seed selection, the weights of all active seeds are normalized to obtain their sampling probabilities.
Consequently, deeply mutated or frequently selected seeds gradually receive lower selection priority, encouraging broader exploration while limiting repeated mutations and unintended semantic deviations.
\section{Experimental Setup}
\label{sec:exmperiment}

To systematically evaluate \tech{}, we investigate the following three research questions:
\begin{itemize}[leftmargin=*]
    \item \textbf{RQ3:} How effective is \tech{} in detecting robustness failures?
    \item \textbf{RQ4:} How do the core components contribute to the effectiveness of \tech{}?
    \item \textbf{RQ5:} How does \tech{} perform in human and LLM-assisted evaluations?
\end{itemize}

\smallskip
\noindent
\textbf{Studied DR Systems.}
We select two representative open-source ReAct-based DR frameworks, WebSailor~\cite{li2025websailor} and WebExplorer~\cite{liu2025webexplorer}, for evaluating \tech{}.
Although DeepDiver~\cite{shi2025deepdiveradaptivesearchintensity} is included in our empirical study, it is excluded from fuzzing experiments because it is a proprietary industrial system.
Its execution trajectories were provided by our industrial collaborators, but we lack the executable access required to inject mutations and resume intermediate executions.
Both selected frameworks support long-horizon web research and have been considered in subsequent studies of web research agents~\cite{wu2025resum,zhang2026argus,pan2026contextpilot}.
Moreover, their open-source implementations and model checkpoints enable us to inspect execution trajectories, modify intermediate search inputs, and resume execution from selected mutation points, which are essential for stage-targeted robustness testing.
WebSailor trains web agents through difficult task synthesis, trajectory-based cold-start training, and agentic reinforcement learning, while WebExplorer employs model-based exploration, iterative query evolution, supervised fine-tuning, and reinforcement learning.
We use their largest officially released model checkpoints for the two frameworks, namely WebSailor-32B\footnote{\url{https://huggingface.co/Alibaba-NLP/WebSailor-32B}} and WebExplorer-8B\footnote{\url{https://huggingface.co/hkust-nlp/WebExplorer-8B}}.
Rather than restricting evaluation to their native framework--model pairings, we combine each framework with both backbone models, yielding four configurations.
This crossed design enables us to examine the effectiveness of \tech{} across different execution frameworks and backbone models.
Together with the four datasets retained for fuzzing, these configurations produce 16 experimental settings.

\begin{table}[t]
    \centering
    \caption{Initial seed selection statistics across framework--model configurations.}
    \label{tab:dataset}
    \begin{adjustbox}{max width=\linewidth,center}
    \small
    \setlength{\tabcolsep}{3pt}
    \begin{threeparttable}
    \begin{tabular}{lrrrrrr}
        \toprule
        \multicolumn{1}{c}{Model} & \multicolumn{1}{c}{Overall} & \multicolumn{1}{c}{FRAMES} & \multicolumn{1}{c}{FRAMES-ZH} & \multicolumn{1}{c}{BrowseComp} & \multicolumn{1}{c}{BrowseComp-ZH} & \multicolumn{1}{c}{OpenSeeker} \\
        \midrule
        Candidates & 4,109 & 824 & 230 & 1,266 & 289 & 1,500 \\
        \midrule
        \multicolumn{7}{@{}l}{\textbf{WebSailor Framework}} \\
        \addlinespace[2pt]
        WebSailor-32B & 775 (18.86\%) & 400 (48.54\%) & 139 (60.43\%) & 90 (7.11\%) & 19 (6.57\%) & 127 (8.47\%) \\
        WebExplorer-8B & 602 (14.65\%) & 366 (44.42\%) & 103 (44.78\%) & 57 (4.50\%) & 13 (4.50\%) & 63 (4.20\%) \\
        \midrule
        \multicolumn{7}{@{}l}{\textbf{WebExplorer Framework}} \\
        \addlinespace[2pt]
        WebSailor-32B & 732 (17.81\%) & 377 (45.75\%) & 130 (56.52\%) & 92 (7.27\%) & 19 (6.57\%) & 114 (7.60\%) \\
        WebExplorer-8B & 271 (6.60\%) & 153 (18.57\%) & 41 (17.83\%) & 28 (2.21\%) & 8 (2.77\%) & 41 (2.73\%) \\
        \bottomrule
    \end{tabular}

    \begin{tablenotes}[flushleft]
    \scriptsize
    \item[] Candidates denotes the number of questions in each benchmark.
    For each framework--model configuration, cells report the number
    of initial seeds answered correctly in all three independent runs
    and their percentage relative to the benchmark size.
\end{tablenotes}
    \end{threeparttable}
    \end{adjustbox}
\end{table}

\smallskip
\noindent
\textbf{Datasets and Initial Seeds.}
We collect candidate seed questions from five datasets: FRAMES~\cite{krishna2024frames}, FRAMES-ZH~\cite{shi2025deepdiveradaptivesearchintensity}, BrowseComp~\cite{wei2025browsecomp}, BrowseComp-ZH~\cite{zhou2025browsecompzh}, and OpenSeeker~\cite{du2026openseeker}.
These datasets cover diverse English and Chinese information-seeking tasks, ranging from multi-hop reasoning to challenging open-web retrieval.
Each question is associated with a reference answer for correctness assessment.
Specifically, FRAMES contains 824 multi-hop questions requiring evidence integration across multiple sources, while FRAMES-ZH provides 230 Chinese-translated questions adopted by DeepDiver.
BrowseComp consists of 1,266 challenging English web-search questions, whereas BrowseComp-ZH contains 289 Chinese questions spanning 11 domains.
OpenSeeker provides synthesized multi-hop question--answer pairs constructed through fact-grounded web-graph expansion and entity obfuscation.
We randomly sample 1,500 question--answer pairs from its released training split as candidate seeds.
In total, the five datasets provide 4,109 candidate questions for each framework--model configuration.
Following the seed-initialization procedure in Section~\ref{sec:seed_initialization}, we independently execute each candidate question three times under each framework--model configuration.
Only questions answered correctly in all three executions are retained as initial seeds, reducing the influence of execution randomness on subsequent pass-to-fail failure identification.
Each configuration maintains a separate seed pool containing the original questions, reference answers, and corresponding execution trajectories.
Table~\ref{tab:dataset} summarizes the number of retained seeds and their retention rates across the five datasets.
Notably, BrowseComp-ZH yields only 19, 13, 19, and 8 retained seeds across the four configurations.
Given its substantially smaller seed pools than the other datasets, we exclude BrowseComp-ZH from the subsequent fuzzing experiments to avoid comparisons based on very limited initial inputs.

\smallskip
\noindent
\textbf{Baselines.}
To evaluate \tech{}, we select two representative approaches with complementary test-generation strategies: RobustAlpacaEval~\cite{cao2024worstprompt} and DR-Arena~\cite{gao2026drarena}.
RobustAlpacaEval generates semantically equivalent paraphrases to evaluate robustness against linguistic variations, whereas DR-Arena dynamically constructs research tasks using live web information to explore DR systems' capability boundaries.
These approaches represent general-purpose input perturbation and DR-specific dynamic task generation, respectively.
Other DR robustness and security studies~\cite{dong2025safesearch,nie2026drnoise,zhu2026misknowagent,pan2026forge,bi2026haegeo} are excluded as baselines because they primarily investigate failures induced by manipulated web content or external evidence, which differ from our input-level intervention setting.
For fair comparisons, all approaches share the same initial seed pool, target framework--model configuration, correctness oracle, and wall-clock testing budget within each experimental setting.
Under this shared testing budget, we execute each generated test input only once, prioritizing broader test exploration over repeated executions given the high cost of long-horizon DR tasks.

\smallskip
\noindent
\textbf{Evaluation Metrics.}
Following prior fuzzing and QA testing studies~\cite{DBLP:conf/icse/YouWCLL23,shen2022qaqa,liu2025cq2a}, we primarily measure testing effectiveness by the number of failure-inducing test inputs discovered within a fixed time budget.
Since \tech{} aims to continuously explore robustness failures rather than measure accuracy on a fixed benchmark, this metric captures failure-discovery capability under comparable testing resources~\cite{dong2025safesearch,nie2026drnoise,zhu2026misknowagent}.
Specifically, we report two metrics based on the semantic answer oracle $\mathcal{O}$ defined in Section~\ref{sec:failure_identification}.
\textit{Failure-Inducing Test Inputs ($N_{\mathrm{fail}}$)} counts generated inputs for which the target DR system produces an incorrect answer relative to the original reference answer, despite consistently answering the corresponding initial seed correctly.
\textit{Passing Test Inputs ($N_{\mathrm{pass}}$)} counts generated inputs for which the target system continues to produce a correct answer.
We report both metrics to characterize failure discovery and the overall volume of evaluated test inputs.
All approaches are evaluated under identical wall-clock budgets within each setting.
A higher $N_{\mathrm{fail}}$ indicates greater failure-discovery capability under the allocated budget, but does not necessarily correspond to a larger number of distinct underlying defects.
Moreover, oracle-identified failures may include invalid mutations or incorrect correctness judgments.
We further assess mutation fidelity and answer-assessment reliability through human and LLM-assisted evaluations in RQ5.

\smallskip
\noindent
\textbf{Implementation.}
We implement \tech{} in Python and employ DeepSeek-V3.2~\cite{deepseekai2025deepseekv32} through an OpenAI-compatible API for mutation generation and semantic answer oracle $\mathcal{O}$.
For mutation generation, we enable thinking mode and adopt the model’s default temperature setting.
The mutation operator and DR systems access Google Search through Serper~\cite{serper} for online fact verification, retrieving up to three search results per query and performing additional searches when necessary.
Web pages are parsed via Jina~\cite{jina} for content extraction.
Following prior fuzzing research~\cite{DBLP:conf/icse/YouWCLL23}, each fuzzing run is allocated a wall-clock budget of 24 hours.
Also, for the seed probability update process, we set $P_{\mathrm{init}}$ to be 1 and $P_{\mathrm{finish}}$ to be 0.05, $\mathrm{max\_times}$ to be 10, $w_1$ to be 1, and $w_2$ to be 0.5.
We record mutation histories, retrieved evidence, execution trajectories, final answers, and oracle judgments to facilitate subsequent analysis and reproducibility.

\section{Results}
\label{sec:results}

\begin{table*}[t]
    \centering
    \caption{Overall performance of \tech{} and baselines.}
    \label{tab:overall_results}
    \small
    \setlength{\tabcolsep}{3pt}
    \begin{adjustbox}{max width=0.9\textwidth,center}
    \begin{threeparttable}
    \begin{tabular}{ll*{10}{r}}
        \toprule
        \multicolumn{1}{c}{\multirow{2}{*}{Approach}} & \multicolumn{1}{c}{\multirow{2}{*}{Model}} &
        \multicolumn{2}{c}{Overall} &
        \multicolumn{2}{c}{FRAMES} &
        \multicolumn{2}{c}{FRAMES-ZH} &
        \multicolumn{2}{c}{BrowseComp} &
        \multicolumn{2}{c}{OpenSeeker} \\
        \cmidrule(lr){3-4}
        \cmidrule(lr){5-6}
        \cmidrule(lr){7-8}
        \cmidrule(lr){9-10}
        \cmidrule(lr){11-12}
        & & Fail & Correct & Fail & Correct & Fail & Correct & Fail & Correct & Fail & Correct \\
        \midrule

        \multicolumn{12}{@{}l}{\textbf{WebSailor Framework}} \\
\addlinespace[1pt]
        \multirow{2}{*}{RobustAlpacaEval}
        & WebSailor-32B & 432 & 1789 & 110 & 776 & 36 & 265 & 160 & 447 & 126 & 301 \\
        & WebExplorer-8B & 309 & 1118 & 82 & 527 & 23 & 159 & 103 & 270 & 101 & 162 \\

        \addlinespace[2pt]

        \multirow{2}{*}{DR-Arena}
        & WebSailor-32B & 895 & 1587 & 255 & 397 & 137 & 288 & 222 & 400 & \textbf{281} & 502 \\
        & WebExplorer-8B & 534 & 1153 & 162 & 285 & 143 & 296 & 110 & 260 & 119 & 312 \\

        \addlinespace[2pt]

        \multirow{2}{*}{\tech{}}
        & WebSailor-32B & \textbf{1429} & 4005 & \textbf{372} & 1034 & \textbf{278} & 1271 & \textbf{502} & 731 & 277 & 969 \\
        & WebExplorer-8B & \textbf{1501} & 4022 & \textbf{317} & 922 & \textbf{367} & 1216 & \textbf{412} & 985 & \textbf{405} & 899 \\

        \midrule
        \multicolumn{12}{@{}l}{\textbf{WebExplorer Framework}} \\
\addlinespace[1pt]
        \multirow{2}{*}{RobustAlpacaEval}
        & WebSailor-32B & 245 & 949 & 41 & 312 & 38 & 210 & 82 & 196 & 84 & 231 \\
        & WebExplorer-8B & 74 & 446 & 23 & 149 & 4 & 37 & 21 & 138 & 26 & 122 \\

        \addlinespace[2pt]

        \multirow{2}{*}{DR-Arena}
        & WebSailor-32B & 727 & 1308 & 247 & 449 & 98 & 164 & 136 & 270 & 246 & 425 \\
        & WebExplorer-8B & 439 & 1024 & 245 & 601 & 75 & 153 & 34 & 107 & 85 & 163 \\

        \addlinespace[2pt]

        \multirow{2}{*}{\tech{}}
        & WebSailor-32B & \textbf{1922} & 4108 & \textbf{579} & 1079 & \textbf{376} & 1286 & \textbf{641} & 722 & \textbf{326} & 1021 \\
        & WebExplorer-8B & \textbf{3488} & 3377 & \textbf{929} & 952 & \textbf{927} & 1027 & \textbf{881} & 655 & \textbf{751} & 743 \\

        \bottomrule
    \end{tabular}

    \begin{tablenotes}[flushleft]
        \footnotesize
        \item[] Fail and Correct denote
        $N_{\mathrm{fail}}$ and $N_{\mathrm{pass}}$,
        respectively.
    \end{tablenotes}

    \end{threeparttable}
    \end{adjustbox}
\end{table*}

\subsection{RQ3: Comparison with Baselines}
\label{sec:rq3}

\noindent
\textbf{Analysis.}
To evaluate the effectiveness of \tech{} in detecting robustness failures, we compare it with RobustAlpacaEval and DR-Arena across 16 framework--model--dataset settings.
For fair comparison, all approaches use the same initial seed pools and a 24-hour testing budget within each setting.
We report the number of failure-inducing (\emph{Fail}) and correctly answered (\emph{Correct}) test inputs to assess failure discovery and test generation capability.
Additionally, we examine the distribution of failures across initial seeds in a representative setting and analyze the failure counts and failure-inducing rates (FI Rates) of the nine mutation rules across all 16 settings.
The FI Rate is calculated as
$N_{\mathrm{fail}}/(N_{\mathrm{fail}}+N_{\mathrm{pass}})$.

\smallskip
\noindent
\textbf{Results.}
Table~\ref{tab:overall_results} presents the overall comparison results across frameworks, models, and datasets.
\ul{In terms of overall failure detection}, \tech{} consistently identifies more failure-inducing inputs than both baselines across all four framework--model configurations.
Specifically, \tech{} identifies 8,340 failure-inducing inputs across the 16 settings, achieving 3.21 and 7.87 times the failure counts of DR-Arena and RobustAlpacaEval, respectively.
At the dataset level, \tech{} achieves the highest failure count in 15 of the 16 settings.
The only exception is on OpenSeeker under the WebSailor framework with WebSailor-32B, where \tech{} identifies slightly fewer failures than DR-Arena, while generating far more correctly answered inputs.
These results indicate that defect-guided fuzzing achieves greater failure-discovery effectiveness than general-purpose input perturbation and dynamic research-task generation under the same testing budget.
However, as these counts rely on automated correctness judgments, they represent potential rather than independently verified robustness failures, whose input fidelity and assessment reliability are further examined in RQ5.

\ul{In terms of framework--model generality}, \tech{} outperforms the strongest baseline, DR-Arena, across all four framework--model configurations, with relative improvements ranging from 59.66\% to 694.53\%.
The largest improvement is observed under the WebExplorer framework with WebExplorer-8B, where \tech{} identifies 7.95 times as many failure-inducing inputs.
Importantly, the improvements are observed in both native and crossed framework--model pairings, suggesting that the effectiveness of \tech{} is not restricted to a particular framework or backbone model among those evaluated.
\ul{In terms of test generation capability}, \tech{} evaluates substantially more inputs within the same testing budget.
Specifically, it generates 3.11 and 4.45 times as many test inputs as DR-Arena and RobustAlpacaEval, respectively.
For correctly answered inputs, the corresponding improvements are 3.06 and 3.61 times.
This advantage is particularly valuable for iterative fuzzing, where promising correctly answered inputs can serve as candidates for further mutation and exploration.
These findings indicate that \tech{} supports more extensive test exploration within a fixed time budget, contributing to its greater failure-discovery capability.

\begin{figure*}[t]
  \centering
  \includegraphics[width=1\linewidth]{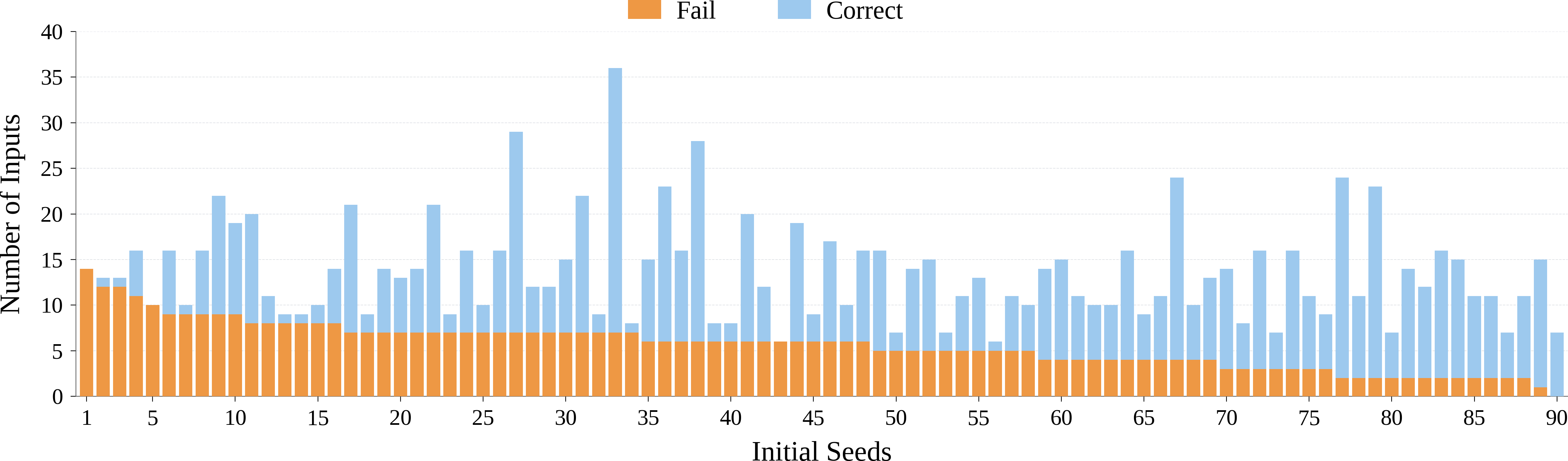}
  \caption{Distribution of failure-inducing and correctly answered inputs across 90 initial seeds on BrowseComp under the WebSailor framework with WebSailor-32B.} 
  \label{fig:seed_results}
\end{figure*}

\ul{In terms of initial-seed coverage}, we examine how failure-inducing inputs are distributed across the 90 initial seeds on BrowseComp under the WebSailor framework with WebSailor-32B.
As shown in Fig.~\ref{fig:seed_results}, \tech{} generates 1,233 test inputs, of which 502 induce potential robustness failures.
Notably, 89 of the 90 initial seeds contribute at least one failure-inducing input, indicating that failure discovery is distributed across a broad range of original questions rather than concentrated in a small subset.
Although failures originating from different seeds do not necessarily correspond to distinct underlying defects, their broad distribution supports the use of initial-seed identity as a lightweight indicator of static diversity, as introduced in Section~\ref{sec:diversity}.

\begin{figure*}[t]
  \centering
  \includegraphics[width=1\linewidth]{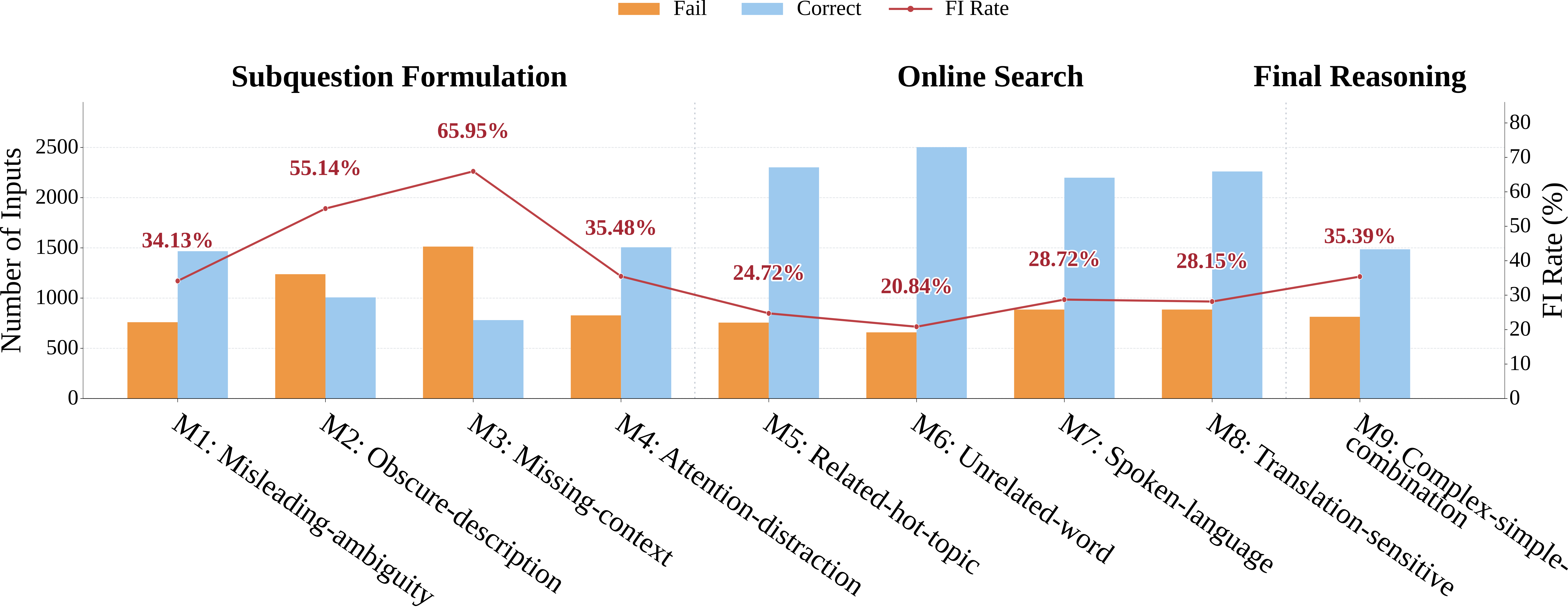}
  \caption{Failure discovery and FI rates of nine defect-guided mutation rules across 16 experimental settings.} 
  \label{fig:mutation_results}
\end{figure*}

\ul{In terms of mutation-rule effectiveness}, Fig.~\ref{fig:mutation_results} presents the failure counts, correctly answered inputs, and FI Rates of the nine defect-guided mutation rules (M1--M9).
All nine rules identify failure-inducing inputs, suggesting that the defect characteristics identified in our empirical study provide useful guidance for the construction of effective robustness tests.
Among them, M3 (Missing-context mutation) and M2 (Obscure-description mutation) achieve the highest failure counts and FI Rates of 65.95\% and 55.14\%, respectively.
Their strong performance suggests that mutations affecting entity identification and contextual constraints are particularly effective at exposing weaknesses during subquestion formulation.
Such interpretation errors may subsequently propagate through online search and final reasoning, consistent with the defect characteristics D2 and D3 identified in our empirical study.
Among online-search mutations, M7 (Spoken-language mutation) and M8 (Translation-sensitive mutation) achieve higher failure counts and FI Rates than M5 and M6.
In contrast, M6 (Unrelated-word mutation) evaluates the largest number of inputs but exhibits the lowest FI Rate (20.84\%).
These results suggest that variations in query expressions and multilingual representations may be more effective at exposing search-related weaknesses than simply introducing irrelevant terms.
Furthermore, M9 (Complex-simple-combination mutation) achieves an FI Rate of 35.39\%, indicating that incorporating simple operations into complex research tasks can also expose potential reasoning weaknesses.
Overall, the nine mutation rules exhibit different failure-discovery patterns across the three execution stages.
These differences provide empirical support for using mutation-rule identity as a complementary indicator of dynamic diversity in Section~\ref{sec:diversity}.
Although different rules do not necessarily reveal distinct underlying defects, their varying effectiveness suggests that they exercise different aspects of DR system robustness and motivate diversity-aware exploration.

\finding{\textbf{Answering RQ3.}
Under the same testing budget, \tech{} achieves 3.21 times the failure-discovery throughput of the strongest baseline, with consistent improvements across framework--model configurations.
Failures are distributed across nearly all examined initial seeds, while all nine mutation rules expose potential robustness failures with varying effectiveness, supporting the use of seed identity and mutation-rule identity as complementary diversity indicators.}

\begin{table}[t]
    \centering
    \caption{Performances of \tech{} and variants across two native framework--model configurations.}
    \label{tab:ablation_results}
    \begin{adjustbox}{max width=0.9\linewidth,center}
    \small
    \setlength{\tabcolsep}{4pt}
    \begin{tabular}{llrrrr}
        \toprule
        \multicolumn{1}{c}{\multirow{2}{*}{Approach}} & \multicolumn{1}{c}{\multirow{2}{*}{Model}} &
        \multicolumn{2}{c}{FRAMES} & \multicolumn{2}{c}{FRAMES-ZH} \\
        \cmidrule(lr){3-4}\cmidrule(lr){5-6}
        & & Fail & Correct & Fail & Correct \\
        \midrule
        \multicolumn{6}{@{}l}{\textbf{WebSailor Framework}} \\
        \addlinespace[4pt]
        \tech{} & WebSailor-32B & \textbf{372} & 1034 & \textbf{278} & 1271 \\
        \textit{w/o dual-stage selection} & WebSailor-32B & 194 & 1312 & 111 & 1338 \\
        \textit{w/o seed maintenance} & WebSailor-32B & 343 & 1056 & 255 & 1190 \\
        \textit{w/o mutation online} & WebSailor-32B & 237 & 796 & 183 & 858 \\
        \midrule
        \multicolumn{6}{@{}l}{\textbf{WebExplorer Framework}} \\
        \addlinespace[4pt]
        \tech{} & WebExplorer-8B & \textbf{929} & 952 & \textbf{927} & 1027 \\
        \textit{w/o dual-stage selection} & WebExplorer-8B & 852 & 1033 & 850 & 1150 \\
        \textit{w/o seed maintenance} & WebExplorer-8B & 903 & 989 & 890 & 1014 \\
        \textit{w/o mutation online} & WebExplorer-8B & 697 & 518 & 691 & 579 \\
        \bottomrule
    \end{tabular}
    \end{adjustbox}

\end{table}

\subsection{RQ4: Ablation Study}
\label{sec:rq4}

\noindent
\textbf{Analysis.}
To investigate the contribution of each core component in \tech{}, we conduct an ablation study by removing one component at a time.
Since DR systems involve long-horizon reasoning and iterative web interactions, repeatedly executing them across numerous configurations incurs substantial computational and time overhead.
To balance evaluation cost and representativeness, we select two native framework--model configurations, WebSailor with WebSailor-32B and WebExplorer with WebExplorer-8B, covering both studied DR frameworks.
We further conduct experiments on FRAMES and FRAMES-ZH, which cover English and Chinese multi-hop research tasks, respectively.
This design allows us to examine component effectiveness across different frameworks and linguistic settings while keeping repeated component-level evaluations tractable.
Specifically, we construct the following three variants:
\begin{itemize}[leftmargin=10pt,nosep]
    \item \textit{w/o dual-stage selection} replaces the dual-stage mutation rule selection strategy with random rule selection, examining the contribution of separating mutation target selection from rule selection.

    \item \textit{w/o seed maintenance} disables potential input evaluation and seed probability updates while preserving all non-failing inputs, probing diversity-aware seed retention and selection.

    \item \textit{w/o mutation online} disables web search during mutation while retaining the DR system's web-search capabilities, examining the contribution of external evidence to the mutation process.
\end{itemize}
All variants use the same initial seed pools, correctness oracle, and 24-hour testing budget as the complete \tech{}.
Following RQ3, we evaluate their effectiveness using the numbers of failure-inducing (\emph{Fail}) and correctly answered (\emph{Correct}) test inputs.

\smallskip
\noindent
\textbf{Results.}
Table~\ref{tab:ablation_results} presents the ablation results across the four experimental settings.
Removing any of the three components consistently reduces the number of failure-inducing inputs compared with the complete \tech{}.
Specifically, removing mutation-time online search, dual-stage selection, and seed maintenance decreases the aggregate failure counts by 27.85\%, 19.91\%, and 4.59\%, respectively, indicating that all three components contribute to failure discovery.
\ul{In terms of web-search-assisted mutation}, removing mutation online search leads to the largest reduction in failure discovery (27.85\%).
This variant also evaluates 32.86\% fewer test inputs, indicating that external search during mutation supports more extensive test generation and exploration.
One possible explanation is that external evidence helps the mutation operator construct and verify modifications involving obscure entities, contextual relationships, and specialized knowledge, thereby facilitating the generation of further test inputs.

\ul{In terms of dual-stage mutation selection}, removing this component reduces failure discovery by 19.91\%, despite evaluating a comparable number of test inputs.
Notably, this variant produces more correctly answered inputs but fewer failure-inducing inputs than the complete \tech{}.
This contrast suggests that dual-stage selection improves the allocation of mutation opportunities toward failure-inducing tests rather than merely increasing the number of evaluated inputs.
By separating mutation target selection from rule selection, \tech{} avoids allocating testing opportunities solely according to the number of rules associated with each target.
The reduction is more pronounced under WebSailor than WebExplorer, indicating that the benefit of this strategy may vary across framework--model configurations.
\ul{In terms of seed maintenance}, removing this component results in a smaller but consistent reduction in failure discovery (4.59\%) across all four settings.
This result suggests that diversity-aware seed maintenance provides additional benefits for iterative fuzzing.
Specifically, potential input evaluation retains non-failing inputs that extend the explored trajectory-length frontier within each diversity state, while seed probability updates reduce repeated selection of deeply mutated or frequently selected seeds.
Together, these mechanisms encourage broader exploration and support the discovery of additional failures through subsequent mutations.

\finding{\textbf{Answering RQ4.}
All three components contribute to the failure-discovery effectiveness of \tech{}.
Removing mutation-time online search, dual-stage selection, and seed maintenance reduces failure counts by 27.85\%, 19.91\%, and 4.59\%, respectively, highlighting their complementary roles in supporting test generation, mutation allocation, and iterative exploration.}

\begin{table}[t]
    \centering
    \caption{Rubric for Human and LLM-assisted evaluation.}
    \label{tab:rq3_rubric}
    \small
    \renewcommand{\arraystretch}{1.15}
    \setlength{\tabcolsep}{2pt}

    \begin{adjustbox}{max width=0.95\textwidth,center}
    \begin{tabularx}{\linewidth}{>{\raggedright\arraybackslash}p{0.24\linewidth}>{\raggedright\arraybackslash}X}
        \toprule
        \multicolumn{1}{c}{Dimension} & \multicolumn{1}{c}{Description} \\
        \midrule
        \multirow{2}{*}{Intent Preservation (IP)} &
        Whether the generated input preserves the original information need, including the intended entities, relations, and requested result. \\
        \midrule
        {Necessary Information Preservation (NIP)} &
        Whether all conditions necessary to determine the original reference answer remain explicit or recoverable from the system-visible context. \\
        \midrule
        {Answer-Correctness Assessment (AC)} &
        Whether the LLM-based correctness judgment of the system's final answer is accurate according to independent assessment against verified evidence. \\
        \bottomrule
    \end{tabularx}
    \end{adjustbox}
\end{table}

\subsection{RQ5: Human and LLM-assisted Evaluation}
\label{sec:rq5}

\noindent
\textbf{Analysis.}
Although RQ3 demonstrates the failure-discovery effectiveness of \tech{}, its automated results depend on the fidelity of generated inputs and the reliability of LLM-based correctness judgments.
To further assess these aspects, following prior studies~\cite{shen2022qaqa,gao2026drarena,chen-etal-2026-beyond-single}, we conduct a human and LLM-assisted evaluation.
Specifically, we randomly sample 384 generated inputs from each of RobustAlpacaEval, DR-Arena, and \tech{}, yielding 1,152 evaluation instances.
The sample size follows the conventional 95\% confidence level and approximately 5\% margin-of-error guideline.
We adopt stratified sampling across datasets, framework--model configurations, and automated outcomes (\emph {Fail} and \emph {Correct}) to ensure balanced coverage.
For \tech{}, the samples are further distributed as evenly as possible across the nine mutation rules.
Each instance includes the original and generated inputs, reference answer, and system response.
For intermediate-search mutations, the preceding execution context is also provided.

We recruit two evaluators proficient in English and Chinese and familiar with multi-step web question answering.
Before the evaluation, both evaluators receive instructions and calibration examples to establish consistent evaluation criteria.
For each approach, 100 instances are independently assessed by both evaluators, while the remaining 284 are equally divided between them.
Thus, each evaluator annotates 242 instances per approach.
To reduce potential bias, approach names and model identities are concealed during annotation.
The evaluation is conducted across multiple sessions over 14 days.
Following prior studies~\cite{morris-etal-2020-reevaluating,shen2022qaqa,zhu2026misknowagent}, we assess three dimensions, as defined in Table~\ref{tab:rq3_rubric}:
\textbf{Intent Preservation (IP)}, \textbf{Necessary Information Preservation (NIP)}, and \textbf{Answer-Correctness Assessment (AC)}.
Each dimension is assigned a binary label, with 1 indicating that the criterion is satisfied.
The inter-rater agreement, measured using Cohen's kappa, is 0.83, indicating strong agreement.
To complement human evaluation with scalable assessment, we additionally employ GPT-6 to independently evaluate the same 1,152 instances using the corresponding criteria.
The results are reported as the percentages of positive labels for each evaluation dimension and approach.

\begin{table}[t]
    \centering
    \caption{Human and LLM-assisted evaluation results (positive-label percentages).}
    \label{tab:rq3_human_llm_results}
    \small
    \renewcommand{\arraystretch}{1.15}
    \setlength{\tabcolsep}{6pt}

    \begin{adjustbox}{max width=0.85\linewidth,center}
    \begin{tabular}{lcccccc}
        \toprule
        \multicolumn{1}{c}{\multirow{2}{*}{Method}}
        & \multicolumn{2}{c}{IP (\%)}
        & \multicolumn{2}{c}{NIP (\%)}
        & \multicolumn{2}{c}{AC (\%)} \\
        \cmidrule(lr){2-3}
        \cmidrule(lr){4-5}
        \cmidrule(l){6-7}
        & Human & GPT-6
        & Human & GPT-6
        & Human & GPT-6 \\
        \midrule
        RobustAlpacaEval
        & 99.22 & 99.48
        & 99.22 & 99.48
        & 89.32 & 90.10 \\
        DR-Arena
        & 28.13 & 19.53
        & 65.89 & 32.81
        & 58.59 & 59.90 \\
        \tech{}
        & 93.49 & 97.66
        & 88.80 & 90.63
        & 88.02 & 86.20 \\
        \bottomrule
    \end{tabular}
    \end{adjustbox}
\end{table}

\smallskip
\noindent
\textbf{Results.}
Table~\ref{tab:rq3_human_llm_results} presents the human and LLM-assisted evaluation results.
\ul{In terms of input fidelity}, \tech{} achieves high preservation rates for both the original information need and conditions necessary to determine the reference answer.
Specifically, human evaluation reports IP and NIP scores of 93.49\% and 88.80\%, respectively, while GPT-6 reports 97.66\% and 90.63\%.
Compared with DR-Arena, \tech{} achieves improvements of 65.36 and 22.91 percentage points in human-assessed IP and NIP, respectively.
These results indicate that most test inputs generated by \tech{} preserve the necessary factual and semantic constraints despite defect-guided modifications.
Interestingly, RobustAlpacaEval achieves even higher fidelity, exceeding 99\% on both dimensions.
This is consistent with its paraphrase-based design, which primarily modifies linguistic expressions while preserving the original semantics.
In contrast, \tech{} introduces targeted changes to entity descriptions, contextual constraints, information salience, and search expressions to exercise empirically identified weaknesses.
These more substantial modifications make fidelity preservation challenging but enable \tech{} to discover more potential robustness failures, as demonstrated in RQ3.
Nevertheless, the remaining invalid inputs indicate that web-search-assisted fidelity assurance reduces, but does not eliminate, unintended semantic changes.

\ul{In terms of answer-assessment reliability}, human evaluation confirms 88.02\% of the automated correctness judgments made for \tech{}'s sampled inputs, comparable to the 89.32\% observed for RobustAlpacaEval.
These results indicate that the semantic answer oracle produces accurate judgments for most evaluated inputs, although assessment errors remain.
In comparison, DR-Arena achieves a substantially lower AC score of 58.59\%.
Notably, its human-assessed IP score is only 28.13\%, indicating that many generated inputs alter the original information need.
Consequently, some answer mismatches against the original reference answer may arise from changes to the task itself rather than genuine pass-to-fail robustness failures.
This distinction highlights that the failure counts reported in RQ3 represent automatically identified failure indications rather than independently verified robustness failures.
Further human inspection reveals that incorrect automated judgments occur among both correctly answered and failure-inducing inputs, with approximately balanced numbers in the two groups.
These errors indicate that automated evaluation can both overlook genuine failures and incorrectly identify successful executions as failures.
Although the substantial differences observed in RQ3 support the failure-discovery effectiveness of \tech{} under a common testing protocol, independent verification remains essential for confirming the validity of individual robustness failures.

\finding{\textbf{Answering RQ5.}
Human and LLM-assisted evaluations show that \tech{} achieves high input fidelity, with human-assessed IP and NIP scores of 93.49\% and 88.80\%, respectively.
Human assessment also confirms 88.02\% of its automated correctness judgments, indicating that \tech{} provides largely reliable robustness tests while leaving room for further improvements in mutation fidelity and answer verification.}

\section{Discussion}
\label{sec:discussion}

\textbf{Model Improvement with \tech-Generated Data.}
Beyond detecting robustness failures, \tech{} can also provide training data for improving DR systems.
Inspired by DeepDiver's use of answer-correctness feedback to optimize web-research policies~\cite{shi2025deepdiveradaptivesearchintensity}, we investigate whether execution trajectories generated by \tech{} can improve the performance of WebSailor-32B.
Specifically, we adopt a two-stage fine-tuning strategy using trajectories generated from OpenSeeker.
First, we perform supervised fine-tuning (SFT) using Low-Rank Adaptation (LoRA)~\cite{hu2022lora} on trajectories with correct final answers.
We then apply Kahneman--Tversky Optimization (KTO)~\cite{pmlr-v235-ethayarajh24a}, treating trajectories with correct and incorrect final answers as positive and negative examples, respectively.
We evaluate the original and fine-tuned models on FRAMES and FRAMES-ZH using identical WebSailor inference settings, without including either evaluation dataset in training.
Each question is executed three times and considered consistently correct only if all three executions produce correct answers.
Table~\ref{tab:model_repair} reports the number of consistently correct questions and the corresponding percentages.
After fine-tuning, the proportion of consistently correct questions increases from 48.54\% to 54.49\% on FRAMES and from 60.43\% to 69.57\% on FRAMES-ZH, representing improvements of 5.95 and 9.13 percentage points, respectively.
These results demonstrate the potential of \tech{}-generated trajectories to improve DR system performance across different benchmarks and languages.
Nevertheless, further mutation-based evaluation is needed to determine whether these improvements also translate into greater robustness against input variations.

\begin{table}[t]
    \centering
    \caption{Consistent@3 performance of WebSailor-32B before and after fine-tuning with \tech{}-generated data.}
    \label{tab:model_repair}
    \small
    \renewcommand{\arraystretch}{1.15}
    \setlength{\tabcolsep}{6pt}
    \begin{adjustbox}{max width=0.9\linewidth,center}
    \begin{tabular}{lrr}
        \toprule
        \textbf{Model} & \textbf{FRAMES} & \textbf{FRAMES-ZH} \\
        \midrule
        WebSailor-32B
        & 400/824 (48.54\%)
        & 139/230 (60.43\%) \\
        WebSailor-32B (LoRA + KTO)
        & \textbf{449/824 (54.49\%)}
        & \textbf{160/230 (69.57\%)} \\
        \bottomrule
    \end{tabular}
    \end{adjustbox}
\end{table}

\smallskip
\noindent
\textbf{Failure Modes of Low-Fidelity Test Inputs.}
Although RQ5 demonstrates that \tech{} preserves the original information need in most generated inputs, some test inputs still violate essential factual or semantic constraints.
To better understand these limitations, we manually inspect the problematic inputs identified in a preliminary annotation of the 384 sampled \tech{} instances.
Based on their primary failure characteristics, we identify three categories:
\textit{(1) Loss of Disambiguating Information (63.2\%).}
Some test inputs inadvertently remove essential clues when obscuring entity descriptions or contextual constraints, making the intended answer ambiguous.
For example, a missing-context mutation replaces the year and score constraints in a Eurovision Song Contest question with vague descriptions, leaving insufficient information to uniquely identify the target countries.
Such failures arise when the mutation operator treats answer-determining constraints as replaceable background information.
This highlights the need to distinguish mutable descriptions from indispensable information during mutation planning and verification.
\textit{(2) Expansion of Requested Output (31.6\%).}
Some test inputs preserve the original research objective but introduce additional answer requirements, thereby changing the expected output.
For instance, a complex-simple-combination mutation extends a question about a former US vice president's age at death by additionally requesting the sum of its digits.
Although the original question remains answerable, the new requirement changes the reference answer and violates strict answer preservation.
This finding highlights the importance of ensuring that auxiliary reasoning operations do not alter the required final output.
Alternatively, test inputs that change the expected answer would require a separately verified reference answer and a corresponding adjustment to the testing oracle.
\textit{(3) Mutation-Rule Noncompliance (5.1\%).}
In some cases, the mutation operator deviates from the assigned rule and produces content that does not constitute a valid test input.
For example, when instructed to translate a search query about the winner of the 2022 Booker Prize into Chinese, the operator instead generates instructions to search using both English and Chinese keywords and compare the results.
Although related to the original task, these instructions do not represent the requested mutated query.
This highlights the need for explicit validation of mutation-rule compliance and output format before test execution.
Overall, these failures reveal three limitations of the current mutation process: insufficient preservation of answer-determining information, unintended changes to output requirements, and deviations from assigned mutation rules.
Addressing these limitations requires more reliable verification of information sufficiency, answer consistency, and mutation-rule compliance to further improve the validity of defect-guided robustness testing.

\section{Threats to Validity}
\label{sec:threats}

\noindent
\textbf{Internal Validity}.~
One potential threat is benchmark contamination, as the evaluated LLMs may have encountered benchmark questions during pretraining.
To reduce its influence on comparisons, all approaches use the same initial seeds, although potential memorization effects cannot be entirely excluded.
Another threat arises from using DeepSeek-V3.2 for both mutation generation and semantic answer evaluation, which may introduce correlated biases.
To mitigate this concern, we employ a separate reference-based evaluation prompt and assess automated judgments through human evaluation in RQ5.
Finally, stochastic LLM inference and dynamic web retrieval may affect execution outcomes.
To mitigate these effects, we require initial seeds to be answered correctly across three independent executions and maintain consistent experimental configurations.
Given the high cost of long-horizon DR execution, we execute each generated test input only once to prioritize broader exploration under fixed testing budgets.
Although this trade-off aligns with our failure-discovery objective, execution randomness may cause some potential failures to be missed or affect the observed failure counts.

\smallskip
\noindent
\textbf{External Validity}.~
One threat is the limited coverage of DR systems and research tasks.
To mitigate this, we evaluate two representative ReAct-based frameworks with two backbone models across multiple English and Chinese benchmarks, and report detailed results across datasets and framework--model configurations.
Nevertheless, our findings may not generalize to other DR architectures, languages, or tasks beyond consistently answered initial seeds.
Another threat arises from the potentially incomplete coverage of our defect taxonomy, which may bias test generation toward known weaknesses.
To alleviate this concern, \tech{} employs diverse initial seeds and multiple defect-guided mutation rules to explore different potential failure behaviors.
However, previously unseen defect patterns may remain uncovered.
Finally, our testing focuses on mutations to user questions and intermediate search inputs, rather than failures caused by webpage accessibility or manipulated web content.
Future work will extend our evaluation to broader DR architectures, defect patterns, and testing scenarios.

\smallskip
\noindent
\textbf{Construct Validity}.~
A primary threat concerns whether automatically identified failure-inducing inputs represent genuine robustness failures.
To mitigate this, \tech{} employs web-search-assisted mutation to preserve input fidelity, while RQ5 evaluates input fidelity and answer-assessment reliability through human and LLM-assisted assessments.
Nevertheless, some generated inputs may alter the original task, particularly in task-generation baselines, and sampled evaluations cannot verify every reported failure.
Another threat concerns failure-origin attribution.
To reduce ambiguity, we follow a consistent annotation procedure that traces the earliest unrecovered errors and their propagation, with annotation criteria refined via independent examination and discussion.
However, without counterfactual interventions, these attributions remain observational rather than causally verified.
Our diversity measurement also uses initial-seed identity and mutation-rule identity as proxies for potential failure diversity.
Although RQ3 examines failure discovery across seeds and mutation rules, different diversity states may not correspond to distinct underlying defects.
Finally, human annotations may introduce subjective judgments.
To mitigate this threat, two evaluators follow predefined criteria and independently assess sampled inputs, achieving a Cohen's kappa coefficient of 0.83, indicating strong inter-rater agreement.

\section{Conclusion}
\label{sec:conclusion}

In this paper, we propose \tech{}, an empirically motivated, defect-guided fuzzing framework for robustness testing of Deep Research systems.
Through an empirical study of DR execution trajectories, we identify nine recurring defect categories across subquestion formulation, online search, and final reasoning, providing guidance for targeted robustness testing.
Building on these findings, \tech{} integrates stage-targeted defect-guided mutation, web-search-assisted fidelity assurance, and diversity-aware seed maintenance to systematically explore potential robustness failures.
Extensive experiments across 16 framework--model--dataset settings show that \tech{} identifies 8,340 failure-inducing test inputs, achieving 3.21 times the failure-discovery throughput of the strongest baseline.
Ablation experiments demonstrate the contributions of its core components, while human and LLM-assisted evaluations confirm that most generated inputs preserve the original task requirements.
Furthermore, experiments using \tech{}-generated data for model fine-tuning improve consistent-answer performance, demonstrating their potential for enhancing DR system reliability.
Future work will explore broader DR system coverage, more reliable mutation validation, and more effective utilization of generated tests for improving model robustness.

\begin{acks}

This work was supported by the National Natural Science Foundation of China under Grant Nos. 62472310, 62322208, 62232001. 

\end{acks}

\bibliographystyle{ACM-Reference-Format}
\bibliography{sample-base}










\end{document}